\documentclass[10pt,twocolumn,letterpaper]{article}

\usepackage[pagenumbers]{cvpr} 

\def\papername{AMB3R}

\usepackage{placeins}
\usepackage[dvipsnames]{xcolor}

\definecolor{firstplace}{rgb}{0.56, 0.78, 0.58}
\definecolor{secondplace}{rgb}{0.9, 1.0, 0.9} 
\definecolor{thirdplace}{rgb}{1.0, 1.0, 0.8} 

\newcommand{\cellfirst}{\cellcolor{firstplace}}
\newcommand{\cellsecond}{\cellcolor{secondplace}}
\newcommand{\cellthird}{\cellcolor{thirdplace}}

\definecolor{papercolor}{rgb}{0.988, 0.549, 0.012}
\newcommand{\papernamecolor}{\textcolor{papercolor}{\textbf{\papername}}}

\usepackage{tikz}
\usetikzlibrary{calc}
\usetikzlibrary{backgrounds}

\usepackage{graphicx}
\usepackage{gensymb}
\usepackage{multirow}
\usepackage{array}
\usepackage{makecell}
\usepackage{tabularx}
\usepackage{booktabs}
\usepackage{colortbl}
\usepackage{pifont}

\usepackage{amsmath}
\usepackage{amssymb}

\usepackage{url}

\usepackage[accsupp]{axessibility} 

\newcommand{\vect}[1]{\boldsymbol{\mathbf{#1}}}

\def\rvu{{\mathbf{i}}}

\def\rvu{{\mathbf{u}}}

\def\gI{{\mathcal{I}}}

\def\gL{{\mathcal{L}}}
\def\gM{{\mathcal{M}}}

\def\gP{{\mathcal{P}}}

\def\gS{{\mathcal{S}}}

\def\gV{{\mathcal{V}}}

\def\sR{{\mathbb{R}}}

\definecolor{cvprblue}{rgb}{0.21,0.49,0.74}
\usepackage[pagebackref,breaklinks,colorlinks,allcolors=cvprblue]{hyperref}

\title{AMB3R: Accurate Feed-forward Metric-scale 3D Reconstruction with Backend}

\author{Hengyi Wang \quad Lourdes Agapito\\
\vspace{1pt}
Department of Computer Science, University College London\\
{\tt\small \url{https://hengyiwang.github.io/projects/amber}
}}

\begin{document}
\twocolumn[{%
    \renewcommand\twocolumn[1][]{#1}%
    \maketitle
    \centering
    \vspace{-0.6cm}
    \vspace{-5pt}
\includegraphics[width=\linewidth]{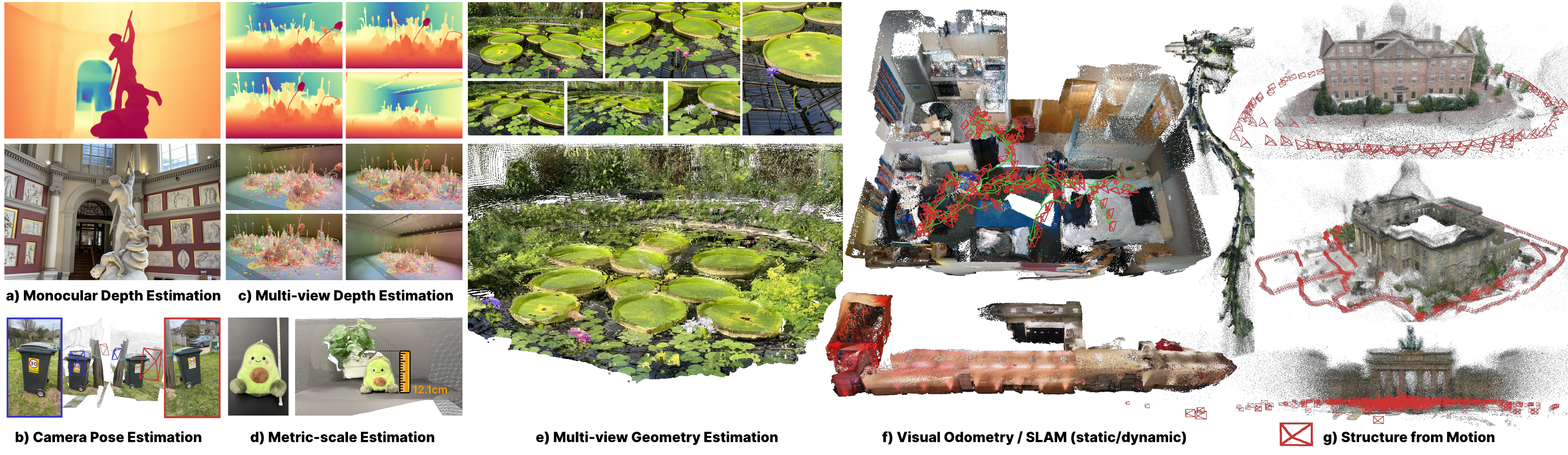}
\vspace{-15pt}
\captionof{figure}{\textbf{Overview.} We present \papername{}, a \textbf{feed-forward model for metric-scale 3D reconstruction}. 
%
%
\papername{} supports camera pose estimation, monocular/multi-view metric depth/3D reconstruction, and can be seamlessly extended to \textbf{visual odometry (VO)/SLAM} and \textbf{Structure from Motion (SfM)}  with \textbf{no task-specific fine-tuning or test-time optimization}. We use in-the-wild images for a)-e), scenes from Co-SLAM~\cite{wang2023co}, TTT3R~\cite{chen2025ttt3r}, and KITTI~\cite{geiger2012kitti} for f), scenes from COLMAP~\cite{schonberger2016colmap}, Tanks\&Temples~\cite{Knapitsch2017tankandtemple}, and IMC PhotoTourism~\cite{jin2021imc} (all images) for g). No confidence threshold is used. f) \& g) results are randomly down-sampled to 3 million points for visualization.}
    \vspace{0.4cm}
}]
\begin{abstract}
We present \papername{}, a multi-view feed-forward model for dense 3D reconstruction on a metric-scale that addresses diverse 3D vision tasks. The key idea is to leverage a sparse, yet compact, volumetric scene representation as our backend, enabling geometric reasoning with spatial compactness. Although trained solely for multi-view reconstruction, we demonstrate that \papername{} can be seamlessly extended to uncalibrated visual odometry (online) or large-scale structure from motion without the need for task-specific fine-tuning or test-time optimization. Compared to prior pointmap-based models, our approach achieves state-of-the-art performance in camera pose, depth, and metric-scale estimation, 3D reconstruction, and even surpasses optimization-based SLAM and SfM methods with dense reconstruction priors on common benchmarks. 
\end{abstract}    
\section{Introduction}
\label{sec:intro}

Pointmaps have emerged as the cornerstone of modern 3D foundation models. By exploiting the one-to-one correspondence between 2D pixels and 3D point coordinates, we can train a neural network to tackle various 3D vision tasks. Yet, this raises a fundamental question: \textit{Is the mapping from 2D pixels to 3D scene points truly one-to-one?}

In practice, this is not the case. Due to visual overlap, multiple pixels often correspond to the same 3D point. This many-to-one mapping, known as \textbf{correspondence}, lies at the heart of decades of research in 3D vision. Early sparse reconstruction methods~\cite{agarwal2009building,davison2007monoslam} leverage correspondences between interest points to recover camera parameters and sparse geometry. Multi-view stereo (MVS) methods~\cite{furukawa2015mvstutorial} construct cost volumes for depth estimation, which essentially match pixel correspondences within the volume. For dense 3D reconstruction, approaches ranging from traditional methods (e.g., KinectFusion~\cite{newcombe2011kinectfusion}) to modern neural implicit models (e.g., DeepSDF~\cite{park2019deepsdf}, NeRF~\cite{mildenhall2020nerf}, NeuralRecon~\cite{sun2021neuralrecon}) adopt compact 3D representations such as TSDF grids, feature grids, or coordinate-based networks. Despite their diverse forms, these representations share a commonality—\textbf{spatial compactness} (i.e., the same 3D coordinate can only have a unique property). This compactness enforces the fusion of multiple corresponding observations of the same scene point into a coherent geometry. 

Recent feed-forward models such as Spann3R~\cite{wang2025spann3r} or VGGT~\cite{wang2025vggt} follow the DUSt3R~\cite{wang2024dust3r} paradigm, formulating the task as per-pixel 2D-to-3D regression. While the pointmap output space implicitly encourages multiple corresponding pixels to have the same 3D location regardless of viewpoint, the network itself operates on 2D grids and lacks explicit geometric reasoning or spatial compactness.

In this paper, we propose \papername{}, a feed-forward reconstruction pipeline with a compact scene representation as backend. We use VGGT as our front-end to predict features and geometry, with an additional scale head trained to recover the metric scale. For the backend, we adopt sparse voxels as our spatial data structure. Thanks to decades of progress in 3D geometry processing, we organize these sparse voxels into a 1D sequence via space-filling curves and process them using a transformer-based architecture, enabling efficient reasoning in compact 3D space. The fused features are injected back into the front-end decoder via zero-convolution layers as in ControlNet~\cite{zhang2023controlnet}, allowing the model to benefit from pre-trained weights and the learned confidence function of the front-end, substantially reducing the training cost ($\sim$ 80 H100 GPU hours).

Furthermore, we demonstrate that \papername{} can be seamlessly extended to perform uncalibrated visual odometry or structure from motion without the need for fine-tuning or test time optimization. To sum up, our contributions are:

\begin{enumerate}
    \item \textbf{A Compact Backend Scene Representation} for a feed-forward pointmap-based foundation model that enables explicit 3D reasoning with spatial compactness.

    \vspace{0.2em}

    \item \textbf{Metric-scale Reconstruction} with a light-weight scale head to recover metric depth information.

    \vspace{0.2em}

    \item \textbf{Uncalibrated Visual Odometry} without the need for fine-tuning or an optimization-based backend. Notably, we show for the first time, feed-forward visual odometry can surpass its optimization-based counterparts.

    \vspace{0.2em}

    \item \textbf{Feed-forward Structure from Motion} capable of large-scale reconstruction without optimization.

    \vspace{0.2em}

    \item \textbf{State-of-the-Art Performance} on 7 tasks across 13 datasets, surpassing existing 3D foundation models and only requiring academic-level training resources ($\sim$ 80 H100 GPU hrs) to train the metric-scale head and new backend on top of frozen VGGT~\cite{wang2025vggt} weights.
    
    \vspace{0.2em}

    \item \textbf{Open Source Release} of code, weights, and an evaluation toolkit to facilitate future research.
\end{enumerate}
\section{Related Work}
\label{sec:formatting}

\subsection{3D Scene Representations}
A 3D scene can be represented through various scene properties -- such as signed distance functions (SDFs), volume density, radiance, surface normals, or learned features -- stored in spatial data structures like voxel grids, point clouds, meshes, or even MLPs. Among these, voxel grids are especially popular in geometric deep learning, as they naturally extend 2D pixels into 3D. Since grids grow cubically with scene size, many sparse alternatives have been introduced, including lossless representations such as voxel hashing~\cite{niessner2013voxelhashing} and octrees~\cite{jackins1980octree1,meagher1982octree2}, as well as lossy ones such as hash feature grids~\cite{muller2022instantngp}, triplanes~\cite{peng2020convoccnet,chan2022eg3d}, and vector–matrix~\cite{chen2022tensorf}, many of which rely on a low-rank approximation of the 3D scene. Point clouds are also a widely used spatial data-structure as they are readily available from posed RGB-D images. Unlike voxels, point clouds are irregular and typically sampled only around visible surface regions. To process such data, HilbertNet~\cite{chen2022hilbertnet} proposes mapping points onto a Hilbert curve~\cite{1blue3brown2017hilbert,hilbert1935stetige} -- a fractal space-filling curve that traverses the entire 3D space while preserving spatial locality -- thereby serializing the point cloud into a 1D sequence that can be efficiently processed with a transformer-based architecture~\cite{zhao2021pointtransformer,wu2022pointtransformerv2,wu2024pointtransformerv3,li2025noksr}. 

\subsection{Pointmap-based Reconstruction}

Pointmaps, or scene coordinates, were originally introduced by SCoRF~\cite{shotton2013scorf} as a scene-specific 3D representation for camera localization~\cite{brachmann2017dsac,brachmann2023accelerated,brachmann2024scene,revaud2024sacreg}. DUSt3R~\cite{wang2024dust3r} and MASt3R~\cite{leroy2024mast3r} extend this concept by training neural networks to regress pointmaps from two-view inputs in a generalizable manner. Since pointmaps inherently encode both camera information and scene geometry, they provide a unified representation that enables a single network to address diverse geometric vision tasks. To scale feed-forward reconstruction beyond two views, Spann3R~\cite{wang2025spann3r} and others~\cite{cabon2025must3r,wang2025cut3r,liu2025slam3r,zhuo2025streamvggt,chen2025long3r,lan2025stream3r,li2025wint3r,wu2025point3r,ma2025puzzles,khafizov2025g,mahdi2025evict3r,chen2025ttt3r} adopt an incremental paradigm in a similar spirit to incremental SfM/SLAM~\cite{davison2007monoslam,schonberger2016colmap}, whereas VGGT~\cite{wang2025vggt} and others~\cite{elflein2025light3rsfm,yang2025fast3r,tang2025mvdust3r,wang2025pi3,wang2025fastervggt} follow a global paradigm that jointly processes all images as in global SfM~\cite{agarwal2009building,pan2024glomap}. Beyond multi-view dense reconstruction, extensions have been proposed for calibrated settings~\cite{jang2025pow3r,keetha2025mapanything}, camera calibration~\cite{lu2024lora3d}, camera localisation~\cite{dong2025reloc3r,huang2025vipe,loiseau2025alligat0r,liu2025regist3r}, monocular depth estimation~\cite{lu2025align3r,li2025lari,sun2025unigeo}, dynamic reconstruction~\cite{jiang2025geo4d,fei2024driv3r,team2025aether,mai2025can,li2025stereodiff,zhang2025pomato}, novel view synthesis~\cite{zhang2025flare,xu2024freesplatter,meuleman2025fly,li2025vicasplat,huang2025video,shi2025revisiting,lin2025movies,lin2025longsplat}, surface reconstruction~\cite{zhu2025surf3r}, semantic segmentation~\cite{zust2025panst3r}, spatial reasoning~\cite{fan2025vlm,hu20253dllm,qian2025gp3}, and scientific imaging~\cite{zhang2025cryofastar}. Since feed-forward approaches are typically constrained to moderate scene sizes~\cite{wang2025spann3r} or limited image counts~\cite{wang2025vggt}, several works~\cite{murai2025mast3rslam,deng2025sail,zhang2025vista,duisterhof2025mast3rsfm,zhou2025mast3rfusion} also explore optimization-based formulations to scale up reconstruction.
\begin{figure*}[t]
    \centering
    \includegraphics[width=\textwidth]{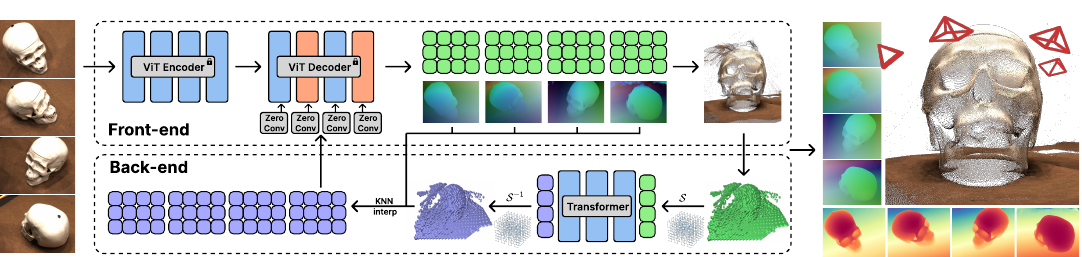}
     \vspace{-18pt}
    \caption{\textbf{Overview of \papername{}.} \papername{} consists of a front-end that predicts pointmaps and geometric features, and a back-end that fuses them into sparse voxels, which are serialized into a 1D sequence, processed by a transformer, and unserialized back to 3D. Per-pixel features are obtained via KNN interpolation and injected into the frozen front-end via zero-convolution for final prediction.} 
    \label{fig:overview}
    \vspace{-10pt}
\end{figure*}

\section{Preliminaries: Pointmap-based Front-end}
\label{sec:pre}

Given a set of input images $\{I_t\}_{t=1}^T$, pointmap-based models independently encode each image into visual features:
\begin{equation}
    F_t = \mathrm{Encoder}(I_t)
\end{equation}
\noindent
Since the pointmap is defined up to a rigid transformation, we fix an anchor image and assume its camera pose to be the identity. The task becomes to predict a pointmap per image expressed in the anchor coordinate system. To achieve this, the model requires a temporal encoding that (\emph{i}) disentangles visual features across images while being generalizable to many images, and (\emph{ii}) identifies the anchor image to provide a consistent reference frame. Existing models can be categorized into three paradigms:

\noindent
\textbf{(a) Network-based temporal encoding.}  
DUSt3R~\cite{wang2024dust3r} employs two intertwined decoders for two images, where the coordinate system is defined by the reference decoder:
\begin{equation}
    P_1^{(1)} = \mathrm{Decoder}^1(I_1), 
    \quad 
    P_2^{(1)} = \mathrm{Decoder}^2(I_2).
\end{equation}
This design choice satisfies condition (\emph{ii}), but inherently limits its scalability to only two images.

\noindent
\textbf{(b) Causal temporal encoding.} 
Spann3R~\cite{wang2025spann3r} proposes an incremental reconstruction framework, where each incoming frame queries accumulated memory features:
\begin{equation}
    P_t^{(1)} = \mathrm{Decoder}(M_{1:t-1}, F_t).
    \label{eq:typeb}
\end{equation}
\noindent 
This formulation naturally satisfies (\emph{i}) and (\emph{ii}) in a causal setting. By sparsifying memory features, Spann3R can process long sequences without exploding GPU memory. 

\noindent
\textbf{(c) Token-based temporal encoding.}  
VGGT~\cite{wang2025vggt} introduces two tokens for the anchor and other images and uses alternating attention as a generalizable temporal encoding:
\begin{equation}
    \{P_t^{(1)}\}_{t=1}^T = \mathrm{Decoder}\big(F_1 + e_{1}, \{F_t + e_{2}\}_{t=2}^T\big).
    \label{eq:typec}
\end{equation}
This enables offline reconstruction with densely connected attention, leading to superior reconstruction quality.

\medskip
\noindent
\textbf{Coordinate normalization.} Pointmap prediction from images is always up to an unknown scale. To address this during training, DUSt3R and Spann3R normalize both the predicted and ground-truth geometry, whereas VGGT normalizes only the ground-truth geometry and trains the model to predict in this normalized space. This forms a canonical space and generally leads to more stable training.

\medskip
\noindent
\textbf{Confidence-aware supervision.} These models are trained with a confidence loss on the ground-truth geometry:
\begin{equation}
    \gL_{\mathrm{conf}} = \sum_t \sum_{i\in\gI} \left(C_t^i \gL_{\mathrm{reg}}(i) - \alpha \log C_t^i\right),
\end{equation}
where $\gI$ denotes the set of valid pixels. Confidence $C_t$ is learned in a self-supervised way that depends on both the data distribution and the training recipe. Consequently, fine-tuning a pre-trained model on different data distributions using a modified training setup without careful initialization may lead to catastrophic forgetting of the learned confidence, potentially degrading reconstruction performance.

\section{Method}

\subsection{Metric-scale Reconstruction}

VGGT~\cite{wang2025vggt} predicts pointmaps normalized by the median distance across all frames (Sec.~\ref{sec:pre}), suggesting that it might implicitly encode metric cues. Thus, we employ a scale head to recover metric scale from its frozen features.

A simple way is to regress the global scale difference between GT and the prediction via a ROE solver~\cite{wang2025moge}. Since this global information depends on ALL frames, we use all intermediate features after the encoder for regression. In practice, though, we find this approach is difficult to train and prone to overfitting, as the scale difference can vary with different frame combinations or even frame order.

Instead, for each frame we regress the \textbf{metric log depth} of  the pixel with median predicted depth. This captures the intrinsic, per-frame property recoverable from individual encoder features, avoiding a strong dependence on the model's (global) prediction, making the task substantially easier. In addition we include the decoder depth feature as guidance. At inference, we estimate per-frame scales and take their median to align the reconstruction to metric space.

\subsection{Backend}

Given pointmap predictions $\{P_t^{(1)}\}_{t=1}^T$ and corresponding geometric features $\{G_t\}_{t=1}^T$ extracted from visual features and decoder features, we align their resolution and construct a sparse voxel grid $\gV$. Voxel features $\{H_i\}_{i=1}^N$ are obtained by averaging features within the same voxel $\gV_i$:
\begin{equation}
    H_i = \frac{1}{|\gP_i|} \sum_{(t,\rvu) \in \gP_i} G_t[\rvu],
\end{equation}
\noindent
where $\gP_i = \{(t,\rvu) \mid P_t^{(1)}[\rvu] \in \gV_i\}$, $\rvu$ denotes the pixel coordinates. We set the voxel size to 0.01 in normalized space, naturally enabling adaptive voxel resolution depending on the scene scale. The sparse voxel grid is serialized into 1D feature sequences via space-filling curves, processed by a transformer, and un-serialized back to voxel space:
\begin{equation}
    \{\hat{H}_i\}_{i=1}^N = (\gS^{-1} \circ f_{\theta} \circ \gS)\left(\{H_i\}_{i=1}^N\right),
\end{equation}
\noindent
where $\gS$ denotes the serialization and $f_{\theta}$ is implemented via Point Transformer v3~\cite{wu2024pointtransformerv3}, a U-Net-like architecture for efficient geometry processing. We then use K-nearest neighbor interpolation to obtain per-point features:
\begin{equation}
\tilde{G}_t[\rvu] = \mathrm{KNN}\big(P_t^{(1)}[\rvu], \{\hat{H}\}_{i=1}^N\big).
\end{equation}
\noindent
These features are fused back to each decoder layer using zero convolution, allowing the model to reuse the attention and confidence functions learned by the front-end model while substantially reducing training cost. See Fig.~\ref{fig:overview}.

\subsection{Training}

We freeze our front-end model (i.e., VGGT~\cite{wang2025vggt}) and only train our backend. Our loss is the same as VGGT for pointmap, depth, and camera pose (except tracking loss): 
\begin{equation}
    \gL = \gL_{\mathrm{depth}}  + \gL_{\mathrm{pointmap}}  + \gL_{\mathrm{camera}}.
\end{equation}
\noindent
Since our training loss, training data, and pre-processing scripts differ from those in VGGT, the expected canonical scale learned by minimizing $\gL$ may not align with that of VGGT. Directly training on normalized data would therefore force the model to waste capacity compensating for this scale mismatch. To mitigate this, we align the predicted geometry with the normalized ground truth using the ROE~\cite{wang2025moge} solver before supervision. For pointmap prediction, we estimate a single scale factor per sequence, while for depth prediction we relax this constraint and estimate an independent scale per image. Because the front-end (VGGT) is trained to produce geometry in a consistent scale and remains frozen, this strategy allows our backend to focus on refining fine structural details without compromising global consistency. The backend is trained for 40 epochs on a mixture of 12 datasets, each epoch containing 2000 samples—a total of 80K samples, still less than a single epoch of VGGT training data—with 5–16 frames per sample, requiring roughly 50 H100 GPU hours to train.

\subsection{Uncalibrated  Visual Odometry}
\label{sec:uvo}

To extend type-(c) methods into a visual odometry (VO) or SLAM system, a straightforward approach is to sequentially build overlapping submaps and estimate both the relative transformations and scale differences between them using the Kabsch–Umeyama algorithm~\cite{kabsch1976solution,umeyama2002least}. Since such alignment introduces non-negligible errors, existing methods usually rely on an optimisation-based backend~\cite{maggio2025vggtslam}. However, we argue this ignores a strong prior in pointmap-based methods: predictions are always expressed in the reference (first) frame coordinate system up to an unknown (median) scale. That is to say, it is not necessary to estimate the transformation for coordinate system alignment.

\noindent
\textbf{Keyframes as memory.} If we rewrite Eq.~\ref{eq:typec} as
\begin{equation}
\{P_t^{(1)}\}_{t=1}^{T-1}, P_T^{(1)}= \mathrm{Decoder}\big(\gM_{1:T-1}, F_T + e_2\big), 
\label{eq:typec2}
\end{equation}
\noindent
where $\gM_{1:T-1} = \left\{F_1 + e_{1}, \{F_t + e_{2}\}_{t=2}^{T-1}\right\}$. Compared to Eq.~\ref{eq:typeb} under the streaming setting, type-(c) methods can be considered as a special memory network that can also update the previous predictions with non-causal attention. Thus, we can use carefully selected keyframes as our memory, enabling the model to run in a visual odometry mode.

\noindent
\textbf{Keyframe selection.} We can measure the pose distance as in VGGT~\cite{wang2025vggt} and our training frame sampling strategy:
\begin{equation}
    D_{i, j} = \arccos\left(\frac{\operatorname{Tr}(R_j R_i^T) - 1}{2}\right) + \lambda \|\tau'_i - \tau'_j\|_2,
\end{equation}
\noindent
where $R$ and $\tau'$ are rotation and normalized translation vector. We set a minimal pose distance with respect to the poses of existing keyframes $\eta_d=0.15$ and iteratively add the frame with the highest confidence within each mapping window as our new keyframe. These keyframes serve as our memory for the next mapping window of $N_w=8$ new frames. Since our prediction is up to the median scale of all input frames, we estimate the scale of the new window $w$ prediction relative to the global map based on keyframe $k\in KF_w$:
\begin{equation}
    s^{w} = \mathrm{ROE}(P_{k}^{(1)}, P_{k}^{(1), w})
    \label{eq:roe}
\end{equation}
After the scale estimation, we will fuse the new window prediction to the existing map in a running average manner:
\vspace{-1em}
\begin{align}
    P_k &\leftarrow (C_k P_k + C_k^w s^{w} P_k^w) / (C_k + C_k^w), \label{eq:update_p} \\
    m   &\leftarrow (C_k m + C_k^w m^w / s^{w}) / (C_k + C_k^w), \\
    \tau_k &\leftarrow (C_k \tau_k + C_k^w s^{w} \tau_k^w) / (C_k + C_k^w), \label{eq:update_t} \\
    q_k &\leftarrow \operatorname{slerp}(q_k, q_k^w, C_k^w / (C_k + C_k^w)), \\
    C_k &\leftarrow C_k + C_k^w. \label{eq:update_c}
\end{align}
\noindent
where $q_k$ is the quaternion vector of the rotation matrix. $C_k$ is the predicted confidence. $m$ is the metric-scale factor.

\noindent
\textbf{Active keyframe management.} Once the number of active keyframes reaches $N_{max}=10$, we resample $N_{min}=7$ keyframes from all historical keyframes. The last keyframe will always be preserved, and we will sample $N_{topk}$ keyframes that have the least pose distance to the last keyframe. To encourage the model to use earlier frames to form a loop, we set a backward search window $\eta_b=0.4$ and iteratively add the earliest keyframe that fits this search window, resulting in $N_{old}$ in total. A keyframe that has the minimal sum pose distance to existing keyframes is sampled to serve as the first frame. Note that once the max pose distance of the active keyframes are larger than $\eta_{max}=1.2$, we resample keyframes to ensure the model only covers a reasonable scene region within the training distribution.

\noindent
\textbf{Coordinate alignment.}
After keyframe resampling, the first keyframe in memory might not be the first frame, resulting in a different coordinate system between our global map and local prediction. Instead of jointly estimating scale and relative transformation from the local map to the global map, we can first map our global map into the local map:
\begin{equation}
    P_k^{(k_0)} = T_{k_0}^{-1}P_k^{(1)},
    \label{eq:align}
\end{equation}
\noindent
where $T_{k_0}=[R_{k_0}, \tau_{k_0}]$. We can then estimate the scale between $P_k^{(k_0)}$ and $P_k^{(k_0), w}$. After scaling the camera pose, we can obtain the relative pose of each corresponding keyframe and use their weighted average to map the local map back to the global frame of reference and fuse them as in Eq.~\ref{eq:update_p}-\ref{eq:update_c}, thus avoiding explicit Kabsch–Umeyama alignment.

\noindent
\textbf{Robust estimation.} To reduce inference cost without sacrificing performance, we only run the backend when the confidence of the front-end prediction falls below a pre-defined threshold. Based on both the confidence and a self-consistency score from the raw pointmap and the unprojected pointmap from the front-end and backend, we either use backend prediction or blend them, ensuring robust performance in the sequential system.

\subsection{Structure from Motion}
Feed-forward pointmap-based models, while efficient, are typically limited by the scale and motion range observed during training. Moreover, type-(c) approaches exhibit quadratic computational complexity with respect to the number of images, making them impractical for large-scale reconstruction. To extend \papername{} into a scalable feed-forward SfM system, we adopt a divide-and-conquer strategy: images are partitioned into small clusters and solved following  incremental SfM without optimization.

\noindent
\textbf{Image clustering.} For each given image $I_t$, we extract its feature descriptor $\bar{F}_t\in\sR^C$, apply feature whitening, and construct the feature distance matrix $D^F\in\sR^{T\times T}$. Furthest Point Sampling (FPS) with iterative splitting and merging is then applied to construct image clusters, each containing between $N_{c_{min}}$ and $N_{c_{max}}$ images.

\noindent
\textbf{Coarse registration.} We initialize the map using the cluster with highest confidence and maintain a global keyframe list with $\eta_d = 0.2$. After initialization, the map is incrementally built by:
(1) identifying the top-k=5 unmapped clusters with the smallest feature distance to existing keyframes, (2) mapping with existing keyframes, and 3) updating global map using the cluster with highest confidence. When the number of global keyframes exceeds $N_{k_{max}} = 8$, we partition the keyframe list into sub-clusters based on pose distance. 
Subsequent  clusters are then mapped against the top-k=5 smaller sub-clusters to improve registration accuracy. 
For each frame, only the prediction from the highest-confidence mapping is kept; low-confidence frames are marked as unregistered and mapped again after registration finishes. Map update and coordinate alignment follow Sec.~\ref{sec:uvo}.

\noindent
\textbf{Global mapping.} After coarse registration, we perform a two-stage mapping as refinement. 
First, we refine all keyframes: starting from the first, we find its top-k closest keyframes within $\eta_r = 1.5$ and perform mapping. We then apply a confidence-prioritized breadth-first search (BFS), adding keyframes in the current mapping window to a priority queue sorted by confidence, thereby traversing the entire keyframe graph, and mapping each keyframe with its closest top-k keyframes. Finally, each non-keyframe is refined by mapping with the top-k closest frames. 

\section{Experiments}

\begin{table}[t!]
\centering
\setlength{\tabcolsep}{2pt}
\resizebox{1.0\columnwidth}{!}{
\begin{tabular}{l cc cc cc cc cc}
\toprule
    \multirow{2}{*}{\textbf{Method}}
    & \multicolumn{2}{c}{\bf NYUv2 \cite{silberman2012nyuv2}}
    & \multicolumn{2}{c}{\bf KITTI \cite{geiger2012kitti}}
    & \multicolumn{2}{c}{\bf ETH3D \cite{schops2017eth3d}}
    & \multicolumn{2}{c}{\bf ScanNet \cite{dai2017scannet}}
    & \multicolumn{2}{c}{\bf DIODE \cite{vasiljevic2019diode}} \\
    
\cmidrule(lr){2-3} \cmidrule(lr){4-5} \cmidrule(lr){6-7} \cmidrule(lr){8-9} \cmidrule(lr){10-11}

& Rel & $\delta_{1.25}$ 
& Rel &$\delta_{1.25}$
& Rel & $\delta_{1.25}$ 
& Rel &$\delta_{1.25}$
& Rel & $\delta_{1.25}$ 
\\

\midrule



Omnidata~\cite{eftekhar2021omnidata} &
7.4 &
94.5 &
14.9 &
83.5 &
16.6 &
77.8 &
7.5 &
93.6 &
33.9 &
74.2
\\


DepthAny v2 \cite{yang2024depthanythingv2} &
\cellthird{4.3} &
\cellthird{97.9} & 
\cellthird{8.0} &
\cellthird{94.4} & 
5.7 &
\cellsecond{98.3} & 
\cellthird{(4.2)} &
(97.9) & 
\cellfirst{21.6} &
75.2
\\

Marigold \cite{ke2023marigold} &
5.5 &
96.4 & 
9.9 &
91.6 & 
6.5 &
96.0 & 
6.4 &
95.1 & 
30.8 &
77.3 \\

Diffusion-E2E\cite{garcia2025diffusione2e} &
5.2 & 96.6 &
9.6 & 91.9 &
6.2 & 95.9 &
5.8 & 96.2 &
30.2 & 77.9 \\

MoGe~\cite{wang2025moge} &
\cellsecond{3.6} & \cellsecond{98.0} &
\cellfirst{5.5} & \cellfirst{97.6} &
\cellsecond{3.3} & \cellfirst{98.8} &
\cellsecond{3.5} & \cellthird{98.3} &
\cellsecond{22.4} & \cellfirst{82.3} \\

\midrule

VGGT~\cite{wang2025vggt} & \cellsecond{3.6} & \cellsecond{98.0} & 8.8 & 92.7 & \cellthird{3.8} & \cellthird{97.9} & \cellfirst{(2.7)} & \cellsecond{(98.8)} & 26.9 & \cellthird{79.1}\\

\papernamecolor{} & \cellfirst{3.0} & \cellfirst{98.9} & \cellsecond{7.3} & \cellsecond{95.4} & \cellfirst{3.2}& \cellfirst{98.8} & \cellfirst{(2.7)} & \cellfirst{(98.9)} & \cellthird{24.7} & \cellsecond{80.7}\\

\bottomrule
\end{tabular}}
\vspace{-6pt}
\caption{\textbf{Monocular depth estimation.} Our method achieves competitive zero-shot monocular depth performance.}
\label{tab:mono_depth}
\vspace{-6pt}
\end{table}


\begin{table}[t!]
    \centering
    \small
    \setlength{\tabcolsep}{2pt}
    \resizebox{\columnwidth}{!}{%
    \begin{tabular}{l|cccccc}
    \toprule
    & \bf DUSt3R~\cite{wang2024dust3r} &  \bf MASt3R~\cite{leroy2024mast3r} & \bf Fast3R~\cite{yang2025fast3r}  & \color{lightgray}VGGT~\cite{wang2025vggt} & \bf VGGT~\cite{wang2025vggt} & \papernamecolor{}\\
    \midrule
    Re10K~\cite{zhou2018realestate10k} & 67.7 & 76.4 & 72.7 & \color{lightgray}85.3 & \underline{81.8} & \textbf{86.3} \\
    \bottomrule
    \end{tabular}
    }%
    \vspace{-6pt}
    \caption{%
    \textbf{Camera Pose Estimation.} We evaluate AUC@30 by randomly sampling 10 frames from the entire sequence.
    }\label{tab:camera_pose}%
    \normalsize
    \vspace{-8pt}
\end{table}

\subsection{Setup}

\noindent\textbf{Benchmarks.} We evaluate across 7 tasks on 13 datasets to show the effectiveness and versatility of our approach:

\begin{itemize}
    \item \textbf{Camera Pose Estimation.} We evaluate against SOTA pointmap-based methods on RealEstate10K~\cite{zhou2018realestate10k}.

    \item \textbf{Monocular Depth Estimation.} We follow prior works~\cite{ke2023marigold} and show zero-shot generalization of our model on five benchmarks: NYUv2~\cite{silberman2012nyuv2}, KITTI~\cite{geiger2012kitti}, ETH3D~\cite{schops2017eth3d}, ScanNet~\cite{dai2017scannet}, and DIODE~\cite{vasiljevic2019diode}.

    \item \textbf{Multi-View (Metric) Depth Estimation.} We compare against various (metric) depth estimation methods on RMVDB~\cite{schroppel2022rmvd}, using KITTI~\cite{geiger2012kitti}, ETH3D~\cite{schops2017eth3d}, ScanNet~\cite{dai2017scannet}, DTU~\cite{aanaes2016dtu}, and Tanks \& Temples~\cite{Knapitsch2017tankandtemple} datasets.

    \item \textbf{3D Reconstruction.} We benchmark against pointmap-based models on ETH3D~\cite{schops2017eth3d} and DTU~\cite{aanaes2016dtu} (RMVDB splits), as well as 7Scenes~\cite{shotton2013scorf} (Spann3R~\cite{wang2025spann3r} splits). Note that Replica~\cite{straub2019replica} is within DUSt3R~\cite{wang2024dust3r} and VGGT~\cite{wang2025vggt} training data, and we remove NRGBD~\cite{azinovic2022nrgbd} due to the potential overlap with synthetic training data.

    \item \textbf{Video Depth Estimation.} We evaluate on Sintel~\cite{butler2012sintel}, BONN~\cite{palazzolo2019bonn}, and KITTI~\cite{geiger2012kitti} without finetuning.

    \item \textbf{VO/SLAM.} Our model is compared against traditional sparse/dense and uncalibrated VO/SLAM methods on static and dynamic scenes from TUM~\cite{sturm2012tumrgbd} and ETH3D SLAM~\cite{schops2019badslam} and pseudo groundtruth on 7scenes~\cite{shotton2013scene}.
    
    \item \textbf{SfM.} We compare against SOTA optimization-based SfM on ETH3D~\cite{schops2017eth3d} and Tanks\&Temples~\cite{Knapitsch2017tankandtemple} datasets.

\end{itemize}

\noindent
\textbf{Metrics.} For depth estimation, we report absolute relative error (rel) and inlier ratios of 3\% ($\delta_{1.03}$) and 25\% ($\delta_{1.25}$). For 3D reconstruction, we evaluate rel ratio as well as accuracy, and completeness, which measures distances between prediction and ground-truth. We remove ICP alignment in Spann3R~\cite{wang2025spann3r} evaluation as pointmap has exact correspondence with ground-truth in contrast to prior works~\cite{wang2023co,zhu2022nice} that requires randomly sampling points from their mesh. We use median-scale alignment for all depth \& 3D reconstruction tasks except metric and monocular depth estimation. For all VO/SLAM tasks, we evaluate the trajectory using ATE RMSE with evo package~\cite{grupp2017evo}. All distance metrics are reported with same unit [cm].

\subsection{Monocular depth estimation}
We compared our zero-shot monocular depth estimation performance in Tab.~\ref{tab:mono_depth} with VGGT~\cite{wang2025vggt} and various representative works. Our model shows consistent improvement over VGGT and SOTA performance on NYUv2~\cite{silberman2012nyuv2} and ETH3D~\cite{schops2017eth3d}, outperforming models specifically trained for monocular depth estimation on those datasets.

\definecolor{bgcolor}{RGB}{190, 181, 190}
\definecolor{mylightgray}{RGB}{238,238,238} 
\colorlet{bgcolor}{mylightgray}
\newcommand{\absrel}{\textrm{rel}}
\newcommand{\threshI}{\delta_{1.03}}
\newcommand{\my}{\ding{51}}
\newcommand{\mn}{\ding{55}}
\newcommand{\bestresult}[1]{\textbf{#1}}
\newcommand{\kittishort}{KITTI}
\newcommand{\scannetshort}{ScanNet}
\newcommand{\ethdshort}{ETH3D}
\newcommand{\dtushort}{DTU}
\newcommand{\tanksandtemplesshort}{T\&T}
\definecolor{mylightgray}{RGB}{238,238,238} 
\colorlet{bgcolor}{mylightgray}
\newcommand{\kitti}{KITTI}
\newcommand{\scannet}{ScanNet}
\newcommand{\dtu}{DTU}
\newcommand{\trainedsimto}[1]{(#1)}
\newcommand{\baselinename}{Robust MVD Baseline}
\newcommand{\absrelname}{Absolute Relative Error}
\newcommand{\otherview}{\other{} view}
\newcommand{\otherviews}{\otherview{}s}
\newcommand{\threshIname}{Inlier Ratio}%
\newcommand{\other}{source}

\begin{table}[t!]
\centering
\setlength{\tabcolsep}{2pt}
\resizebox{\columnwidth}{!}{
\begin{tabular}{|l|c >{\columncolor{bgcolor}} c
|c >{\columncolor{bgcolor}} c
|c >{\columncolor{bgcolor}} c
|c >{\columncolor{bgcolor}} c
|c >{\columncolor{bgcolor}} c
|c >{\columncolor{bgcolor}} c
|}
\hline
 \textbf{Approach}
& \multicolumn{2}{c|}{\textbf{\kittishort{}}}
& \multicolumn{2}{c|}{\textbf{\scannetshort{}}}
& \multicolumn{2}{c|}{\textbf{\ethdshort{}}}
& \multicolumn{2}{c|}{\textbf{\dtushort{}}}
& \multicolumn{2}{c|}{\textbf{\tanksandtemplesshort{}}}
& \multicolumn{2}{c|}{\textbf{Average}}
\\

& $\absrel\downarrow$ & $\threshI\uparrow$
& $\absrel\downarrow$ & $\threshI\uparrow$
& $\absrel\downarrow$ & $\threshI\uparrow$
& $\absrel\downarrow$ & $\threshI\uparrow$
& $\absrel\downarrow$ & $\threshI\uparrow$
& $\absrel\downarrow$ & $\threshI\uparrow$
\\
\hline
\hline

\multicolumn{13}{|l|}{\textbf{a) Classic approaches}}
\\
COLMAP~\cite{schonberger2016colmap}
	& 12.0
	& 58.2
	& 14.6
	& 34.2
	& 16.4
	& 55.1
	& \cellfirst{0.7}
	& \cellfirst{96.5}
	& 2.7
	& \cellfirst{95.0}
	& 9.3
	& 67.8
	\\

\hline
\hline
\multicolumn{13}{|l|}{\textbf{b) Single-view depth}}
\\

UniDepthV2~\cite{piccinelli2025unidepthv2}
&4.0
&55.3
&\cellthird{(2.1)}
&\cellthird{(82.6)}
&3.7
&66.2
&3.2
&72.3
&3.6
&68.4
&3.3
&68.9
\\

	DepthAny V2~\cite{yang2024depthanythingv2}
	& 6.6
	& 38.6
	& 4.0
	& 58.6
	& 4.7
	& 56.5
	& 2.6
	& 74.7
	& 4.5
	& 57.5
	& 4.8
	& 54.1
	\\ %

\hline
\hline

\multicolumn{13}{|l|}{\textbf{c) Depth from frames and poses (w/ per-image range)}}
\\

	Vis-MVSNet~\cite{zhang2023vismvsnet}
	& {9.5}
	& {55.4}
	& 8.9
	& 33.5
	& {10.8}
	& {43.3}
	& (1.8)
	& (87.4)
	& {4.1}
	& {87.2}
	& {7.0}
	& {61.4}
	\\ %

MVSFormer++~\cite{cao2024mvsformerplus}
&{4.4}
&{65.7}
&{7.9}
&{39.4}
&{7.8}
&{50.4}
&\cellsecond{(0.9)}
&\cellthird{(95.3)}
&{3.2}
&{88.1}
&{4.8}
&{67.8}
\\

\hline
\hline

\multicolumn{13}{|l|}{\textbf{d) Depth from frames and poses (w/o per-image range)}}
\\

	Robust MVD \cite{schroppel2022rmvd}
	& {7.1}
	& 41.9
	& {7.4}
	& {38.4}
	& {9.0}
	& {42.6}
	& {2.7}
	& {82.0}
	& {5.0}
	& {75.1}
	& {6.3}
	& {56.0}
	\\ %

MVSA~\cite{izquierdo2025mvsanywhere}%
&\cellthird{3.2}
&\cellthird{68.8}
&3.7
&62.9
&3.2
&68.0
&1.3
&95.0
&\cellthird{2.1}
&\cellsecond{90.5}
&2.7
&77.0
\\

\hline
\hline

\multicolumn{13}{|l|}{\textbf{e) Depth from frames (w/o poses)}}
\\

DeMoN~\cite{ummenhofer2017demon}
& 15.5
& 15.2
& 12.0
& 21.0
& 17.4
& 15.4
& 21.8
& 16.6
& 13.0
& 23.2
& 16.0
& 18.3
\\ %

DUSt3R~\cite{wang2024dust3r} & 5.4 & 49.5 & (3.1) & (71.8) & 3.0 & 76.0 & 3.9 & 68.6 & 3.3 & 75.1 & 3.7 & 68.2 \\
Spann3R~\cite{wang2025spann3r} & 7.9 & 36.2 & (3.3) & (67.1) & 5.7 & 58.6 & 3.5 & 65.2 & 4.7 & 58.5 & 5.0 & 57.1 \\
Pow3R~\cite{jang2025pow3r} & 5.7 & 45.7 & (3.2) & (68.8) & 3.0 & 74.7 & 3.0 & 74.3 & 3.3 & 76.6 & 3.6 & 68.0 \\
MUSt3R~\cite{cabon2025must3r} & 4.5 & 55.0 & (4.0) & (59.8) & 2.5 & 80.3 & 4.6 & 55.4 & (2.6) & (80.4) & 3.7 & 66.2\\

VGGT~\cite{wang2025vggt} & 4.5 & 59.6 & (2.3) & (80.8) & \cellthird{1.8} & \cellthird{86.3} & \cellsecond{0.9} & \cellsecond{95.6} & 2.4 & 84.1 & \cellthird{2.4} & \cellthird{81.3} \\

$\pi$3$^\ddagger$~\cite{wang2025pi3} & \cellfirst{2.8} & \cellsecond{72.9} & \cellsecond{(2.0)} & \cellsecond{(83.6)} & \cellfirst{1.3} & \cellfirst{92.4} & 1.3 & 91.8 & \cellsecond{1.8} & 87.3 & \cellsecond{1.8} & \cellsecond{85.6} \\

MapAny$^\ddagger$~\cite{keetha2025mapanything} & 4.0 & 59.4 & 4.0 & 60.5 & 2.8 & 73.2 & 3.9 & 63.7 & 3.3 & 73.0 & 3.6 & 66.0 \\

\papernamecolor{} & \cellfirst{2.8} & \cellfirst{74.4} & \cellfirst{(1.9)} & \cellfirst{(85.8)} & \cellsecond{1.4} & \cellsecond{90.9} & \cellsecond{0.9} & 95.1 & \cellfirst{1.7} & \cellthird{90.2} & \cellfirst{1.7} & \cellfirst{87.3} \\
\hline
\end{tabular}}
\vspace{-5pt}
\caption{\textbf{Multi-view depth estimation} on RMVDB~\cite{schroppel2022rmvd}. $\ddagger$ means concurrent works, and () indicate training dataset. Please refer to supplement for full table.}
\vspace{-10pt}
\label{tab:mv_depth}
\end{table}
\subsection{Camera pose estimation.}

\begin{figure*}[t]
    \centering
    \includegraphics[width=0.99\textwidth]{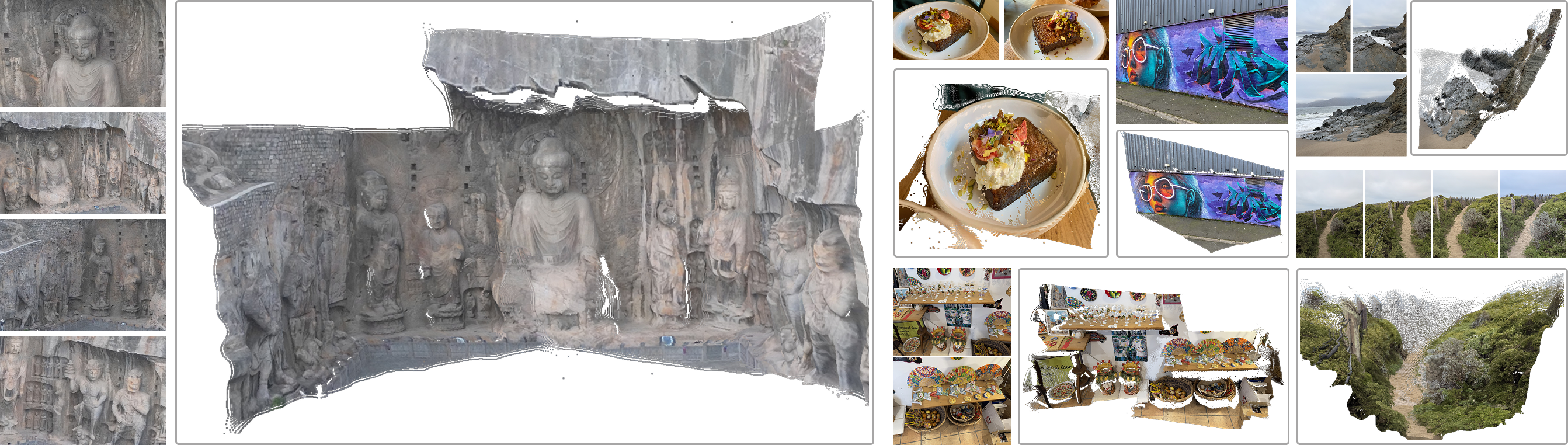}
     \vspace{-5pt}
    \caption{\textbf{Qualitative showcase} of generalization to in-the-wild images such as Longmen Grottoes~\cite{zheng2025culture3d}.} 
    \label{fig:qual}
    \vspace{-10pt}
\end{figure*}

\begin{table}[t!]
\centering
\setlength{\tabcolsep}{1pt}
\resizebox{1.0\columnwidth}{!}{
\begin{tabular}{l ccc ccc ccc ccc ccc}
\toprule
    \multirow{2}{*}{\textbf{Method}}
    & \multicolumn{3}{c}{\textbf{\kittishort{}}~\cite{geiger2012kitti}}
    & \multicolumn{3}{c}{\textbf{\scannetshort{}}~\cite{dai2017scannet}}
    & \multicolumn{3}{c}{\textbf{\ethdshort{}}~\cite{schops2017eth3d}}
    & \multicolumn{3}{c}{\textbf{\dtushort{}}~\cite{aanaes2016dtu}}
    & \multicolumn{3}{c}{\textbf{\tanksandtemplesshort{}}~\cite{Knapitsch2017tankandtemple}}
    \\
    \cmidrule(lr){2-4} \cmidrule(lr){5-7} \cmidrule(lr){8-10} \cmidrule(lr){11-13} \cmidrule(lr){14-16}
    & $\absrel$ & $\delta_{1.03}$ & $\delta_{1.25}$
    & $\absrel$ & $\delta_{1.03}$ & $\delta_{1.25}$
    & $\absrel$ & $\delta_{1.03}$ & $\delta_{1.25}$
    & $\absrel$ & $\delta_{1.03}$ & $\delta_{1.25}$
    & $\absrel$ & $\delta_{1.03}$ & $\delta_{1.25}$
    \\
\midrule
MASt3R~\cite{leroy2024mast3r} &
56.8& 0.1& 8.4&
(10.5)& \cellsecond{(27.0)}& (91.0)&
40.4& 10.3& 26.7&
129.0& 1.7& 8.1&
63.7& 0.0& 1.5\\

MUSt3R~\cite{cabon2025must3r} &
82.4& 0.0& 0.0&
(40.2)& (0.7)& (6.2)&
63.3& 1.0& 3.9&
\cellfirst{46.8}& 2.6& 25.3&
77.0& 0.0& 0.0\\

Spann3R~\cite{wang2025spann3r} &
16.4& \cellthird{18.9}& 76.5&
(33.1)& (3.3)& (23.0)&
25.1& 18.6& 65.3&
\cellsecond{67.8}& \cellthird{8.7}& \cellthird{42.7}&
\cellthird{23.3}& \cellthird{17.6}& \cellthird{57.4}\\

CUT3R~\cite{wang2025cut3r} &
28.6& 4.1& 29.5&
\cellthird{(9.9)}& (19.2)& \cellfirst{(\bf 96.3)}&
29.4& 6.4& 40.7&
138.5& 0.7& 2.3&
46.7& 0.1& 1.9\\

MapAny$^\ddagger$~\cite{keetha2025mapanything} &
\cellthird{12.8}& 16.1& \cellthird{78.2}&
31.9& 7.7& 41.2&
\cellthird{15.4}& \cellthird{26.6}& \cellthird{83.1}&
718.5& 0.0& 0.2&
24.6& 15.3& 41.5\\

\papernamecolor{} (dec. sc.) &
\cellfirst{4.5}& \cellfirst{56.5}& \cellfirst{97.8}&
\cellfirst{(\bf 9.3)}& \cellfirst{( 41.4)}& \cellthird{( 91.7)}&
\cellsecond{12.2}& \cellsecond{42.1}& \cellsecond{83.9}&
\cellthird{96.2}& \cellsecond{25.6}& \cellsecond{59.3}&
\cellsecond{9.4}& \cellsecond{42.6}& \cellsecond{83.8}\\

\papernamecolor{} (enc. z) &
\cellsecond{8.2}& \cellsecond{27.9}& \cellsecond{95.6}&
\cellfirst{(9.3)}& \cellthird{( 26.9)}& \cellsecond{( 95.2)}&
\cellfirst{8.5}& \cellfirst{48.3}& \cellfirst{90.4}&
240.9& \cellfirst{30.3}& \cellfirst{60.9}&
\cellfirst{6.3}& \cellfirst{49.9}& \cellfirst{95.7}\\

\bottomrule
\end{tabular}}
\vspace{-5pt}
\caption{\textbf{Multi-view metric-scale depth estimation} on RMVDB~\cite{schroppel2022rmvd}. $\ddagger$: concurrent work.}
\label{tab:mv_depth_metric}
\vspace{-5pt}
\end{table}

\begin{table}[t!]
\centering
\setlength{\tabcolsep}{5pt}
\resizebox{\columnwidth}{!}{
\begin{tabular}{l l ccc ccc ccc}
\toprule
& \multirow{2}{*}{\textbf{Method}}
& \multicolumn{3}{c}{\textbf{\ethdshort{}}}
& \multicolumn{3}{c}{\textbf{\dtushort{}}}
& \multicolumn{3}{c}{\textbf{7-Scenes}}
\\ \cmidrule(lr){3-5} \cmidrule(lr){6-8} \cmidrule(lr){9-11}
& & $\absrel\!\downarrow$ & Acc$\!\downarrow$ & Cp$\!\downarrow$
& $\absrel\!\downarrow$ & Acc$\!\downarrow$ & Cp$\!\downarrow$
& $\absrel\!\downarrow$ & Acc$\!\downarrow$ & Cp $\!\downarrow$
\\ 

\midrule

\multirow{3}{*}{\rotatebox[origin=c]{90}{Onl.}} &
Spann3R~\cite{wang2025spann3r} &
24.96& 47.63& 31.87 &
4.96& 1.89 & 0.64 &
8.58 & 4.81 & 5.48 \\

& MUSt3R~\cite{cabon2025must3r} &
19.91& 45.19& 82.72 &
5.26& 1.68 & 1.09 &
11.57 & 8.94 & 14.97 \\

& CUT3R~\cite{wang2025cut3r} &
18.83& 38.90& 22.93 &
9.11& 3.75 & 0.96 &
6.32 & 2.88 & \cellsecond{3.26} \\

\midrule

\multirow{4}{*}{\rotatebox[origin=c]{90}{Offl.}} &
VGGT~\cite{wang2025vggt} &
\cellthird{6.02}& \cellthird{12.81}& \cellthird{11.89} &
\cellsecond{0.83}& \cellfirst{0.22} & \cellfirst{0.08} &
\cellsecond{5.51} & \cellsecond{2.32} & 3.51 \\

& $\pi3^\ddagger$~\cite{wang2025pi3} & \cellsecond{5.82}& \cellsecond{10.54}& \cellsecond{10.42} &
\cellthird{1.57}& \cellthird{0.50} & \cellthird{0.29} &
\cellthird{5.92} & \cellthird{2.60} & 3.42 \\

& MapAny$^\ddagger$~\cite{keetha2025mapanything} &
11.20& 21.87& 19.76 &
10.37& 4.35 & 1.54 &
7.61 & 3.48 & \cellthird{3.33} \\

 & \papernamecolor{} &
\cellfirst{4.64}& \cellfirst{9.98}& \cellfirst{9.69} &
\cellfirst{0.81}& \cellfirst{0.22} & \cellfirst{0.08} &
\cellfirst{4.74} & \cellfirst{1.74} & \cellfirst{2.84} \\

\bottomrule
\end{tabular}}
\vspace{-5pt}
\caption{\textbf{Multi-view 3D Reconstruction [cm].} For ETH3D and DTU, we use image tuples from RMVDB~\cite{schroppel2022rmvd}. For 7 scenes, we use frames as in Spann3R with improved GT poses (See Tab.~\ref{tab:slam_7scenes}).}
\label{tab:3d_recon}
\vspace{-10pt}
\end{table}

Following~\cite{wang2024dust3r,wang2025vggt}, we evaluate camera pose estimation (see Tab.~\ref{tab:camera_pose}) by randomly sampling 10 frames per scene and evaluating AUC@30. For VGGT, we report their official results (gray) (which include some non-accessible datasets), and our run of VGGT on datasets available to us. \papername{} outperforms VGGT in both cases with our performance gain resulting purely from the improved geometric features.


\subsection{Multi-view (metric) depth estimation} We conduct a comprehensive evaluation on RMVDB~\cite{schroppel2022rmvd} in Tab.~\ref{tab:mv_depth}, where our model establishes a new SOTA on RMVDB, outperforming concurrent pointmap-based models ($\pi$3~\cite{wang2025pi3} and MapAnything~\cite{keetha2025mapanything}) while requiring one order of magnitude less training resources. Our model also outperforms SOTA depth estimation model MVSA~\cite{izquierdo2025mvsanywhere}, which requires ground-truth poses and intrinsics. 

For metric depth evaluation (Tab.~\ref{tab:mv_depth_metric}), we report results under both 3\% and 25\% inlier thresholds, as 3\% is overly strict for metric-depth methods. Our approach achieves the best overall performance under both metrics. Compared to directly regressing scale-differences from decoder features, our log-depth regression yields overall better results. On DTU, larger absolute errors occur in scenes with toy buildings placed on white tables with dark backgrounds, which some models mistakenly interpret as real structures.

\subsection{3D reconstruction}

We evaluate 3D reconstruction in Tab.~\ref{tab:3d_recon} and show some qualitative examples in Fig.~\ref{fig:qual}. For object-level reconstruction (DTU), both our model and VGGT achieve millimeter accuracy with less than 1\% error. For outdoor (ETH3D) and indoor (7-scenes) scenes, we achieve SOTA performance compared to concurrent work $\pi$3~\cite{wang2025pi3}.

\begin{table}[t!]
\centering
\setlength{\tabcolsep}{5pt} 
\resizebox{\columnwidth}{!}{%
\begin{tabular}{l ccc ccc ccc}
\toprule
\multirow{2}{*}{\textbf{Method}}
& \multicolumn{3}{c}{\textbf{Sintel}}
& \multicolumn{3}{c}{\textbf{Bonn}}
& \multicolumn{3}{c}{\textbf{KITTI}}
\\ \cmidrule(lr){2-4} \cmidrule(lr){5-7} \cmidrule(lr){8-10}
& $\absrel\!\downarrow$ & $\delta_{1.03}\!\uparrow$ & $\delta_{1.25}\!\uparrow$
& $\absrel\!\downarrow$ & $\delta_{1.03}\!\uparrow$ & $\delta_{1.25}\!\uparrow$
& $\absrel\!\downarrow$ & $\delta_{1.03}\!\uparrow$ & $\delta_{1.25}\!\uparrow$
\\ \midrule

Spann3R~\cite{wang2025spann3r} &
55.2& 7.3& 46.3 &
5.5& 40.9& 97.8 &
24.7 &  10.4 & 64.4 \\

CUT3R~\cite{wang2025cut3r} &
58.4& 9.6& 46.6 &
5.1& 47.4& 98.0 &
13.6 &  18.0 & 81.0 \\


VGGT~\cite{wang2025vggt} &
\cellsecond{28.0}& \cellsecond{13.6}& \cellthird{61.8} &
\cellthird{3.6}& \cellthird{63.2}& \cellthird{98.4} &
\cellthird{8.2} &  \cellthird{40.4} & \cellthird{90.9} \\

$\pi^3$~\cite{wang2025pi3} &
\cellthird{31.9}& \cellthird{13.3}& \cellfirst{66.8} &
\cellfirst{3.0}& \cellfirst{72.0} & \cellfirst{98.8} &
\cellfirst{3.6} & \cellfirst{59.5} & \cellfirst{98.8} \\

\papernamecolor{} &
\cellfirst{23.6} & \cellfirst{14.0}&  \cellsecond{62.7} &
\cellsecond{3.3} & \cellsecond{66.0} & \cellsecond{98.5} &
\cellsecond{4.1} & \cellsecond{53.8} & \cellsecond{98.3} \\
\bottomrule
\end{tabular}%
}
\vspace{-5pt}
\caption{\textbf{Video depth estimation on Sintel, Bonn, and KITTI}. We evaluate all methods on all scenes from Sintel and Bonn, rather than selecting 14 and 4 scenes as in prior works. In contrast to Pi3~\cite{wang2025pi3}, we did not do specific finetuning on dynamic datasets.}
\label{tab:video_depth}
\vspace{-4pt}
\end{table}

\newcommand{\lgray}{\color{lightgray}}

\begin{table}[t]
    \centering
    \scriptsize
    \renewcommand\arraystretch{1.1}
    \setlength{\tabcolsep}{4pt}
    \resizebox{\columnwidth}{!}{%
    \begin{tabular}{llcccccccc}
    \toprule
     & & \textbf{chess} & \textbf{fire} & \textbf{head} & \textbf{office} & \textbf{pump} & \textbf{redkit} & \textbf{stair} & \textbf{avg} \\
    \midrule
    & Pseudo GT~\cite{newcombe2011kinectfusion} & 3.5 & 2.5 & \bf1.4 & 9.4 & 15.1 & 4.9 & \bf2.9 & 5.7 \\
    & \papernamecolor{} (ALL) & \bf1.8 & \bf1.8 & 2.1 & \bf4.7 & \bf2.4 & \bf1.6 & 3.7 & \bf2.1 \\

    \midrule
    \multirow{3}{*}{PSNR} & COLMAP (GT)~\cite{brachmann2021newgt7scene} & 25.3 & 25.7 & 22.6 & 25.8 & 25.0 & 21.9 & 24.7 & 24.4 \\
    
    & Pseudo GT~\cite{newcombe2011kinectfusion} & 21.2 & 22.1 & 21.2 & 22.6 & 22.6 & 19.9 & 23.1 & 21.8\\

    & \papernamecolor{} & 20.6 & 22.3 & 20.2 & 22.0 & 23.6 & 21.8 & 22.2 & 21.8\\
    
    \bottomrule
    \end{tabular}%
    }
    \vspace{-5pt}
    \caption{\textbf{VO: ATE RMSE [cm] on 7 scenes}. We show our online model surpasses the pseudo GT prior works used~\cite{wang2025spann3r,murai2025mast3rslam} on ATE and achieve on-par novel view synthesis PSNR. COLMAP~\cite{brachmann2021newgt7scene} GT is obtained via all sequences on each scene.}
    \label{tab:slam_7scenes}
    \vspace{-10pt}
\end{table}
\begin{table}
\begin{center}
\renewcommand\arraystretch{1.2}
\setlength{\tabcolsep}{1pt} 
\tiny
\hspace{-3mm}
\resizebox{1.0\columnwidth}{!}{
\begin{tabular}{cl|ccccccccccc|c}
\specialrule{0.6pt}{0.5pt}{0.5pt}
  & & \multicolumn{8}{c}{fr1} & \multicolumn{2}{|c|}{fr2} & \multicolumn{1}{c|}{fr3} &\\
  & & 360 & desk & desk2 & plant & room & rpy & teddy & xyz & \multicolumn{1}{|c}{xyz} & \multicolumn{1}{c|}{desk} & long & AVG\\
  \specialrule{0.4pt}{0.5pt}{0.5pt}
  \multirow{3}{*}{S} & ORB-SLAM3 \cite{mur2015orb} & X & \cellsecond{2.0} & X & 11.8 & X & 5.6 & X & \cellfirst{1.0} & \cellthird{0.5} & \cellfirst{1.3} & \cellfirst{1.7} & X \\
  & DSO \cite{engel2017dso} & X & 27.2 & 66.0 & 6.0 & 58.6 & X & X & 3.8 & \cellsecond{0.3} & \cellthird{2.2} & 9.9 & X \\
  & DPVO \cite{lipson2024dpvslam} & 13.1 & 9.4 & \cellthird{6.5} & \cellsecond{3.0} & 39.8 & 3.5 & \cellsecond{6.2} & 1.3 & \cellthird{0.5} & 3.5 & 5.5 & \cellthird{8.4} \\
\specialrule{0.4pt}{0.5pt}{0.5pt}
\multirow{7}{*}{D} & TANDEM \cite{koestler2022tandem} & X & 4.3 & 33.7 & X & X & 4.9 & 43.1 & 2.4 & \cellsecond{0.3} & \cellsecond{2.0} & 8.3 & X \\
  & MonoGS \cite{matsuki2024monogs} & 14.2 & 6.3 & 74.0 & 9.3 & 64.9 & 3.4 & 35.6 & 1.6 & 4.5 & 133.1 & \cellsecond{3.3} & 31.8 \\
  & DeepFactors \cite{czarnowski2020deepfactors} & 17.9 & 15.9 & 20.2 & 31.9 & 38.3 & 3.8 & 56.0 & 5.9 & 8.4 & 26.3 & 49.0 & 24.9 \\
  & DepthCov \cite{dexheimer2023depthcov} & \cellthird{12.8} & 5.6 & \cellsecond{4.8} & 26.1 & \cellthird{25.7} & 5.2 & 47.5 & 5.6 & 1.2 & 15.9 & 68.8 & 19.9 \\
  & DROID-VO \cite{teed2021droidslam} & 15.7 & 5.2 & 11.1 & 6.0 & 33.4 & \cellthird{3.2} & 19.1 & 5.6 & 10.7 & 7.9 & 7.3 & 11.4 \\
  & COMO \cite{dexheimer2024como} & 12.9 & 4.9 & 9.5 & 13.8 & 27.0 & 4.8 & 24.5 & 4.0 & 0.7 & 6.3 & 10.5 & 10.8\\
  & GlORIE-VO\cite{zhang2024glorieslam} & 13.1 & \cellthird{4.0} & 8.6 & \cellthird{4.1} & 32.7 & \cellsecond{2.9} & 14.5 & \cellthird{1.2} & \cellfirst{0.2} & 16.1 & 4.8 & 9.3 \\
\specialrule{0.4pt}{0.5pt}{0.5pt}
 & Spann3R \cite{wang2025spann3r} & 20.7 & 16.1 & 28.3 & 57.4 & 84.8 & 6.1 & 92.4 & 2.1 & 4.4 & 20.7 & 193.9 & 47.9\\
U & MUSt3R~\cite{cabon2025must3r} &  \cellsecond{8.9} & 5.1 & 7.1 & 5.4 & \cellsecond{13.4} & 5.2 & \cellthird{6.9} & 2.7 & 1.7 & 15.6 & \cellthird{4.3} & \cellsecond{7.1}\\

  & \lgray MUSt3R$^\star$~\cite{cabon2025must3r} & \lgray 7.8 & \lgray 4.0 & \lgray 4.6 & \lgray 4.0 & \lgray 9.9 & \lgray 4.3 & \lgray 4.2 & \lgray 1.3 & \lgray 1.2 & \lgray 15.3 & \lgray 4.3 & \lgray 5.5 \\

  & \bf\papernamecolor{} (ALL) & \cellfirst{4.6} & \cellfirst{1.9} & \cellfirst{2.8} & \cellfirst{2.9} & \cellfirst{5.8} & \cellfirst{2.3} & \cellfirst{3.7} & \cellsecond{1.1} & 2.1 & 3.4 & 5.0 & \cellfirst{3.2}\\
  & \lgray \papernamecolor{} (KF) & \lgray 3.9 & \lgray 1.7 & \lgray 2.3 & \lgray 2.7 & \lgray 5.5 & \lgray 2.2 & \lgray 2.8 & \lgray 0.8 & \lgray 0.6 & \lgray 3.0 & \lgray 4.0 & \lgray 2.7\\
\specialrule{0.6pt}{0.5pt}{0.5pt}
\end{tabular}
}
\vspace{-5pt}
\caption{\textbf{VO: ATE RMSE [cm] on TUM RGB.} We compare with Sparse (S) versus dense (D) versus dense unconstrained (U) methods. The baseline methods are re-run without Loop Closure and global bundle adjustment. We compare with MUSt3R~\cite{cabon2025must3r} on all frames while keeping keyframe predictions as a reference. }
\label{tab:vo_tum}
\end{center}
\vspace{-15pt}
\end{table}
\begin{table}[t]
\centering
\scriptsize
\setlength{\tabcolsep}{3pt}
\resizebox{\columnwidth}{!}{
\begin{tabular}{l lcccccccccc} 

\toprule
& &\textbf{360} &\textbf{desk} &\textbf{desk2} &\textbf{floor} &\textbf{plant} &\textbf{room } &\textbf{rpy} &\textbf{teddy} &\textbf{xyz} &\textbf{avg} \\
\midrule

\multirow{8}{*}{\rotatebox[origin=c]{90}{\bf Calibrated}} &\textbf{ORB-SLAM3 \cite{mur2015orb}} &X &\cellsecond{1.7} &21.0 &X &3.4 &X &X &X &\cellsecond{0.9} &- \\
&\textbf{DeepV2D \cite{teed2020deepv2d}} &24.3 &16.6 &37.9 &165.3 &20.3 &24.6 &10.5 &31.6 &6.4 &37.5 \\
&\textbf{DeepFactors \cite{czarnowski2020deepfactors}} &15.9 &17.0 &25.3 &16.9 &30.5 &36.4 &4.3 &60.1 &3.5 &23.3 \\
&\textbf{DPV-SLAM \cite{lipson2024dpvslam}} &11.2 &\cellthird{1.8} &\cellthird{2.9} &5.7 &\cellthird{2.1} &33.0 &\cellthird{3.0} &8.4 &\cellthird{1.0} &7.6 \\
&\textbf{DPV-SLAM++ \cite{lipson2024dpvslam}} &13.2 &\cellthird{1.8} &\cellthird{2.9} &5.0 &2.2 &9.6 &3.2 &9.8 &\cellthird{1.0} &5.4 \\
&\textbf{GO-SLAM \cite{zhang2023goslam}} &8.9 &\cellfirst{1.6} &\cellthird{2.8} &\cellsecond{2.5} &2.6 &\cellsecond{5.2} &\cellfirst{1.9} &4.8 &\cellthird{1.0} &\cellthird{3.5} \\
&\textbf{DROID-SLAM \cite{teed2021droidslam}} &11.1 &\cellthird{1.8} &4.2 &\cellfirst{2.1} &\cellfirst{1.6} &\cellfirst{4.9} &2.6 &4.8 &1.2 &3.8 \\
&\textbf{MASt3R-SLAM~\cite{murai2025mast3rslam}} &\cellsecond{4.9} &\cellfirst{1.6} &\cellsecond{2.4} &\cellsecond{2.5} &\cellsecond{2.0} &6.1 &2.7 &\cellthird{4.1} &\cellsecond{0.9} &\cellsecond{3.0} \\
\midrule
\multirow{5}{*}{\rotatebox[origin=c]{90}{\bf Uncalibrated}} &\textbf{DROID-SLAM\cite{teed2021droidslam}} &20.2 &3.2 &9.1 &6.4 &4.5 &91.8 &5.6 &4.5 &1.2 &15.8 \\
&\textbf{MASt3R-SLAM~\cite{murai2025mast3rslam}} &\cellthird{7.0} &3.5 &5.5 &5.6 &3.5 &11.8 &4.1 &11.4 &2.0 &6.0 \\
& \bf VGGT-SLAM~\cite{maggio2025vggtslam} & 7.1& 2.5& 4.0& 14.1 & 2.3& 10.2& \cellthird{3.0} & \cellsecond{3.4}& 1.4& 5.3\\


  & \lgray\papernamecolor{}  (ALL) &  \lgray4.6 & \lgray1.9 & \lgray2.8& \lgray3.2& \lgray2.9& \lgray5.8& \lgray2.3 & \lgray3.7& \lgray1.1 & \lgray3.1\\
 
  & \papernamecolor{} (KF) &  \cellfirst{3.9} & \cellsecond{1.7} & \cellfirst{2.3}& \cellthird{2.7}& 2.7& \cellthird{5.5}& \cellsecond{2.2} & \cellfirst{2.8}& \cellfirst{0.8} & \cellfirst{2.7}\\
\bottomrule
\end{tabular}}
\vspace{-5pt}
\caption{\textbf{SLAM: ATE RMSE [cm] on TUM RGB.} We use image stride of 2 and keyframe poses evaluation as in MASt3R-SLAM~\cite{murai2025mast3rslam} with all poses evaluation shown as a reference. Note that we did not do any post-processing on trajectories }
\label{tab:slam_tum}
\vspace{-10pt}
\end{table}

\begin{table}[ttt]
\begin{center}
\setlength{\tabcolsep}{3pt} 
\resizebox{0.95\columnwidth}{!}{
\begin{tabular}{l ccccccccc c}
\toprule
& \bf cables1 & \bf camshake1 & \bf einstein1 & \bf plant1 & \bf plant2 & \bf sofa1 & \bf table3 & \bf table7 & \bf avg\\
\midrule
Spann3R~\cite{wang2025spann3r} 
              & 33.2 & 5.1 & 30.9 & 4.1 & 5.7 & 17.1 & 19.3 & 18.9 & 16.8 \\ 
MUSt3R~\cite{cabon2025must3r}    & 20.7&5.6&15.4&2.3&2.7&15.8&17.6& 9.5&11.2\\
\lgray MUSt3R$^\star$~\cite{cabon2025must3r}  &\lgray 20.7&\lgray 5.3&\lgray 11.2&\lgray 1.8 & \lgray 2.7& \lgray 15.5&\lgray 17.3 &\lgray 5.5 & \lgray 10.0\\

\papernamecolor{} (ALL) &\bf 2.8 &\bf 2.4&\bf 2.3&\bf 1.5 &\bf 1.6&\bf 3.6&\bf 3.7 & \bf 2.8 & \bf 2.6\\

\lgray\papername{} (KF) &\lgray2.3&\lgray2.1&\lgray1.9&\lgray0.5 &\lgray 0.9&\lgray 3.1&\lgray2.6 &\lgray2.4 & \lgray2.0\\

\bottomrule

\end{tabular}}
\normalsize
\vspace{-6pt}
\caption{\textbf{VO: ATE RMSE [cm] on ETH3D SLAM~\cite{schops2019badslam}.} We use the same scene sets and image stride of 2 as in MUSt3R~\cite{cabon2025must3r}}
\label{tab:vo_eth3d}
\end{center}
\vspace{-20pt}
\end{table}

\begin{table}[t]
\centering
\scriptsize
\setlength{\tabcolsep}{2pt}
\resizebox{0.95\columnwidth}{!}{
\begin{tabular}{lcccccccccc} 

\toprule
&\textbf{sit\_xyz} &\textbf{desk\_p} &\textbf{walk\_rpy} &\textbf{walk\_s} &\textbf{walk\_h} &\textbf{sit\_s} &\textbf{sit\_h} &\textbf{walk\_x} &\textbf{sit\_r} &\textbf{avg} \\
\midrule
\textbf{MUSt3R~\cite{cabon2025must3r}} & 3.5 & 3.3 & 6.7 & 2.2 & 5.5 & 1.5 & 5.3 & 7.8 & 4.9 & 4.5\\
\lgray\textbf{MUSt3R$^\star$~\cite{cabon2025must3r}} & \lgray2.4 & \lgray2.2 & \lgray5.0 & \lgray1.9 & \lgray3.9 & \lgray1.3 & \lgray4.6 & \lgray5.4 & \lgray4.0 & \lgray3.4 \\

\lgray\textbf{MegaSaM~\cite{li2025megasam}} & \lgray1.1 & \lgray OOM & \lgray3.3 & \lgray0.6 & \lgray1.8 & \lgray0.6 & \lgray2.5 & \lgray1.4 & \lgray2.5 & -\\

 \textbf{\papernamecolor{}} & \bf1.3 & \bf2.1 & \bf4.2 & \bf0.9 & \bf1.9 & \bf0.8 & \bf2.2 & \bf2.0 & \bf2.0 & \bf1.9\\

\bottomrule
\end{tabular}}
\vspace{-6pt}
\caption{\textbf{VO: ATE RMSE on TUM Dynamic.} MUSt3R$^\star$uses re-rendering with trajectory smoothness. MegaSaM uses keyframe \& global BA at the end followed by a trajectory smoother.}
\label{tab:vo_tum_dynamic}
\vspace{-8pt}
\end{table}

\subsection{Dynamic reconstruction}

We evaluate video depth estimation in Tab.~\ref{tab:video_depth}. To align with other depth estimations tasks for static scenes, we additionally report inlier ratio of 3\% for a reality check. Despite \papername{} not being specifically trained for dynamic scene reconstruction, it still shows consistent improvement over VGGT~\cite{wang2025vggt} and competitive reconstruction results compared to concurrent work $\pi$3\cite{wang2025pi3}, which includes dynamic scenes at training. This lays a foundation for making a VO system possible in dynamic environments.

\subsection{Uncalibrated Visual Odometry/SLAM}

We present VO/SLAM results in Tab.~\ref{tab:slam_7scenes}-~\ref{tab:vo_tum_dynamic}. Our model, Spann3R~\cite{wang2025spann3r}, and MUSt3R~\cite{cabon2025must3r}, together establish a new class of \emph{uncalibrated visual odometry} methods with unknown camera intrinsics. Both MUSt3R and our approach outperform classic optimization-based VO, with our model further reducing the tracking error of previous SOTA from $7.1$cm to $3.2$cm on TUM and from $11.2$cm to $2.6$cm on ETH3D. Our method even surpasses pseudo GT used in prior work~\cite{wang2025spann3r,murai2025mast3rslam} on 7-Scenes (Tab.~\ref{tab:slam_7scenes}), where we validate ground-truth reliability through novel-view synthesis.

We further compare against state-of-the-art SLAM systems in Tab.~\ref{tab:slam_tum}, where baseline methods are allowed to use loop closure and global bundle adjustment as post-processing after tracking the last frame. The impact of these post-processing steps can be seen by comparing Tab.~\ref{tab:vo_tum} and Tab.~\ref{tab:slam_tum}. Following prior work~\cite{murai2025mast3rslam,maggio2025vggtslam}, we evaluate on raw keyframe poses for fair comparison. Our method shows for the first time an uncalibrated  feed-forward approach outperforming optimization-based methods (with calibration and post-processing) on the TUM~\cite{sturm2012tumrgbd} dataset.


Compared to hybrid systems combining dense priors with optimization-based backends, such as MASt3R-SLAM~\cite{murai2025mast3rslam} and VGGT-SLAM~\cite{maggio2025vggtslam}, our model achieves superior performance in the uncalibrated setting, and even surpasses MASt3R-SLAM with known calibration.

Finally, we evaluate on dynamic environments (Tab.~\ref{tab:vo_tum_dynamic}), where our model again surpasses the previous SOTA uncalibrated VO, MUSt3R~\cite{cabon2025must3r}. Neither model is trained for dynamic reconstruction, yet our raw predictions remain comparable to MegaSaM~\cite{li2025megasam}, a method explicitly designed for dynamic scenes and equipped with global bundle adjustment and trajectory smoothing as post-processing steps. Please refer to suppl. material/video for qualitative results.

\subsection{Structure from Motion}
We report SfM results in Tab.~\ref{tab:sfm}-\ref{tab:sfm_tnt}. Unlike all baselines, our method operates in a feed-forward manner without optimization-based BA. Despite this, it achieves substantial improvements in rotation accuracy over MASt3R-SfM~\cite{duisterhof2025mast3rsfm} on ETH3D~\cite{schops2017eth3d}, an unordered image collection. On Tanks\&Temples~\cite{Knapitsch2017tankandtemple,Riegler2020FVS} with video input, our method further surpass ACE-Zero~\cite{brachmann2024scene} by a large margin. Please refer to suppl. material/video for extensive qualitative results.

\begin{table}[t]
    \centering
    \setlength{\tabcolsep}{1pt}

\resizebox{\columnwidth}{!}{
\begin{tabular}{lr@{~}rr@{~}rr@{~}rr@{~}rr@{~}rr@{~}r}
\toprule
        \multirow{2}{*}{Scenes} & \multicolumn{2}{c}{COLMAP~\cite{schonberger2016colmap}} & \multicolumn{2}{c}{ACE-Zero~\cite{brachmann2024scene}} & \multicolumn{2}{c}{VGGSfM~\cite{wang2025vggt}} & \multicolumn{2}{c}{DF-SfM~\cite{he2024dfsfm}} & \multicolumn{2}{c}{MASt3R-SfM} & \multicolumn{2}{c}{\papernamecolor{} (SfM)} \\
        \cmidrule(lr){2-3} \cmidrule(lr){4-5} \cmidrule(lr){6-7} \cmidrule(lr){8-9} \cmidrule(lr){10-11} \cmidrule(lr){12-13}
                 &  \multicolumn{1}{c}{\footnotesize RRA@5} &  \multicolumn{1}{c}{\footnotesize RTA@5} & \multicolumn{1}{c}{\footnotesize RRA@5} & \multicolumn{1}{c}{\footnotesize RTA@5} &  \multicolumn{1}{c}{\footnotesize RRA@5} &  \multicolumn{1}{c}{\footnotesize RTA@5} & \multicolumn{1}{c}{\footnotesize RRA@5} & \multicolumn{1}{c}{\footnotesize RTA@5} & \multicolumn{1}{c}{\footnotesize RRA@5} & \multicolumn{1}{c}{\footnotesize RTA@5} & \multicolumn{1}{c}{\footnotesize RRA@5} & \multicolumn{1}{c}{\footnotesize RTA@5}  \\
\midrule
{\small courtyard} & {56.3} & {60.0} & {4.0} & {1.9} & {50.5} & {51.2} & \cellthird{80.7} & \cellsecond{74.8} & \cellsecond{89.8} & \cellthird{64.4} & \cellfirst{100.0} & \cellfirst{96.5} \\
{\small delivery area} & {34.0} & {28.1} & {27.4} & {1.9} & {22.0} & {19.6} & \cellthird{82.5} & \cellfirst{82.0} & \cellsecond{83.1} & \cellsecond{81.8} & \cellfirst{91.0} & \cellthird{76.6} \\
{\small electro} & {53.3} & {48.5} & {16.9} & {7.9} & {79.9} & {58.6} & \cellthird{82.8} & \cellsecond{81.2} & \cellfirst{100.0} & \cellfirst{95.5} & \cellsecond{95.6} & \cellsecond{81.2} \\
{\small facade} & \cellsecond{92.2} & \cellsecond{90.0} & {74.5} & {64.1} & {57.5} & {48.7} & \cellthird{80.9} & \cellthird{82.6} & {74.3} & {75.3} & \cellfirst{100.0} & \cellfirst{95.4} \\
{\small kicker} & {87.3} & {86.2} & {26.2} & {16.8} & \cellfirst{100.0} & \cellthird{97.8} & {93.5} & {91.0} & \cellfirst{100.0} & \cellfirst{100.0} & \cellfirst{100.0} & \cellsecond{99.2} \\
{\small meadow} & {0.9} & {0.9} & {3.8} & {0.9} & \cellfirst{100.0} & \cellfirst{96.2} & \cellthird{56.2} & \cellthird{58.1} & \cellsecond{58.1} & \cellthird{58.1} & \cellfirst{100.0} & \cellsecond{95.2} \\
{\small office} & {36.9} & {32.3} & {0.9} & {0.0} & \cellthird{64.9} & {42.1} & \cellsecond{71.1} & \cellsecond{54.5} & \cellfirst{100.0} & \cellfirst{98.5} & \cellfirst{100.0} & \cellthird{53.9} \\
{\small pipes} & {30.8} & {28.6} & {9.9} & {1.1} & \cellfirst{100.0} & \cellsecond{97.8} & {72.5} & {61.5} & \cellfirst{100.0} & \cellfirst{100.0} & \cellfirst{100.0} & \cellthird{87.9} \\
{\small playground} & {17.2} & {18.1} & {3.8} & {2.6} & {37.3} & {40.8} & \cellthird{70.5} & \cellsecond{70.1} & \cellfirst{100.0} & \cellfirst{93.6} & \cellsecond{98.7} & \cellthird{62.2} \\
{\small relief} & {16.8} & {16.8} & {16.8} & {17.0} & \cellsecond{59.6} & \cellsecond{57.9} & {32.9} & {32.9} & \cellthird{34.2} & \cellthird{40.2} & \cellfirst{100.0} & \cellfirst{90.1} \\
{\small relief 2} & {11.8} & {11.8} & {7.3} & {5.6} & \cellsecond{69.9} & \cellthird{70.3} & {40.9} & {39.1} & \cellthird{57.4} & \cellfirst{76.1} & \cellfirst{100.0} & \cellsecond{75.7} \\
{\small terrace} & \cellfirst{100.0} & \cellfirst{100.0} & {5.5} & {2.0} & {38.7} & {29.6} & \cellfirst{100.0} & \cellsecond{99.6} & \cellfirst{100.0} & \cellfirst{100.0} & \cellfirst{100.0} & \cellthird{97.2} \\
{\small terrains} & \cellfirst{100.0} & \cellfirst{99.5} & {15.8} & {4.5} & \cellthird{70.4} & \cellthird{54.9} & \cellfirst{100.0} & \cellsecond{91.9} & {58.2} & {52.5} & \cellsecond{91.6} & {53.8} \\
\midrule
Average & {49.0} & {47.8} & {16.4} & {9.7} & {65.4} & {58.9} & \cellthird{74.2} & \cellthird{70.7} & \cellsecond{81.2} & \cellsecond{79.7} & \cellfirst{98.2} & \cellfirst{81.9} \\
\bottomrule
\end{tabular}}
\vspace{-5pt}
\caption{%
    \textbf{SfM on ETH3D~\cite{schops2017eth3d}.} Our method does not need optimization-based BA, unlike all baseline methods.
    }\label{tab:sfm}
    \vspace{-10pt}
\end{table}
\begin{table}[t!]
\centering
\setlength{\tabcolsep}{5pt}
\resizebox{\columnwidth}{!}{
\begin{tabular}{l cc cc cc} 
\toprule
\multirow{2}{*}{\textbf{Method}}
& \multicolumn{2}{c}{\textbf{Training} (7 scenes)}
& \multicolumn{2}{c}{\textbf{Intermediate} (8 scenes)}
& \multicolumn{2}{c}{\textbf{Advanced} (6 scenes)}
\\ 

\cmidrule(lr){2-3} \cmidrule(lr){4-5} \cmidrule(lr){6-7}
& RRA@5 & RTA@5 
& RRA@5 & RTA@5 
& RRA@5 & RTA@5
\\ \midrule

\bf ACE-Zero~\cite{brachmann2024scene} &
\cellsecond{73.9}& \cellsecond{72.9} &
\cellsecond{67.6}& \cellsecond{74.0} &
\cellthird{22.9} & \cellthird{19.1} \\

\bf MASt3R-SfM~\cite{duisterhof2025mast3rsfm} &
\cellthird{56.2}& \cellthird{64.9} &
\cellthird{50.8}& \cellthird{57.5} &
\cellsecond{38.8} & \cellsecond{36.5} \\

\papernamecolor{} (SfM) &
\cellfirst{95.0}& \cellfirst{94.5} &
\cellfirst{98.7} & \cellfirst{96.9}  &
\cellfirst{68.0} & \cellfirst{72.4}  \\

\bottomrule
\end{tabular}}
\vspace{-5pt}
\caption{\textbf{SfM on Tanks\&Temples~\cite{Knapitsch2017tankandtemple}.} Each scene contains 151 to
1106 images. We only retain methods succeed on every scene.}
\label{tab:sfm_tnt}
\vspace{-5pt}
\end{table}

\begin{figure}[t]
    \centering
    \includegraphics[width=\columnwidth]{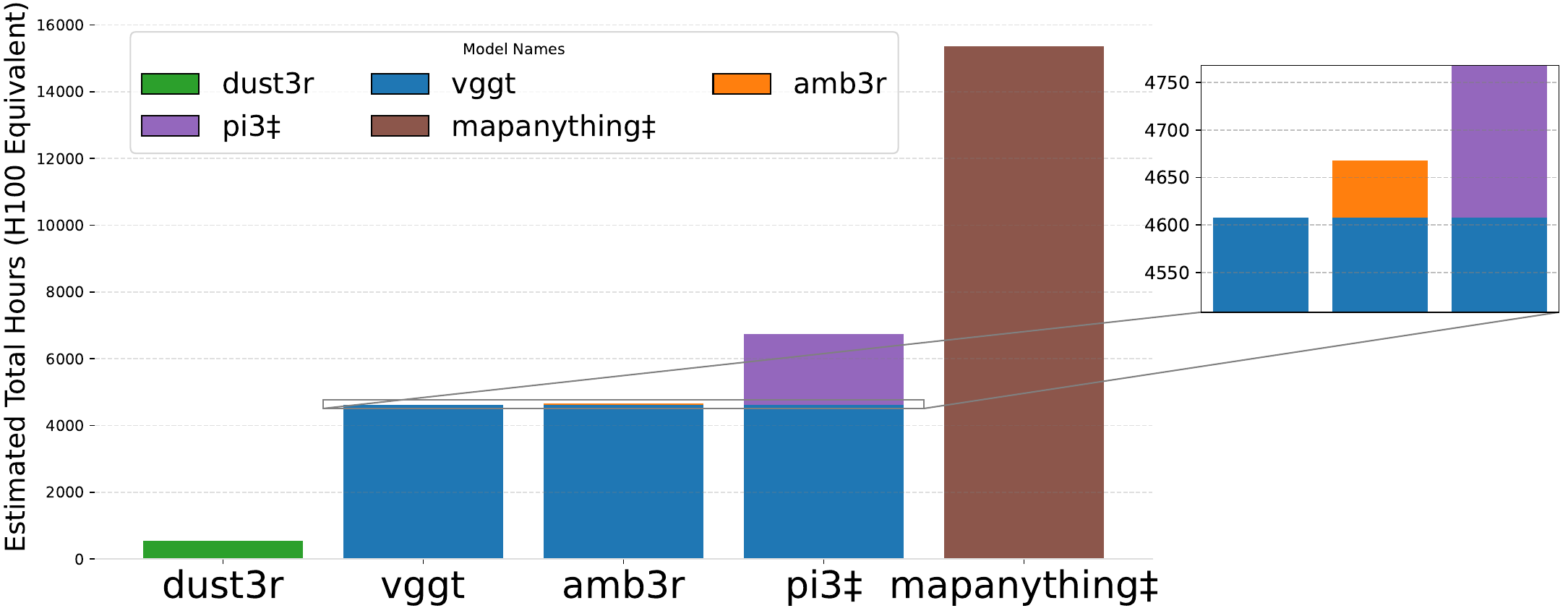}
     \vspace{-18pt}
    \caption{\textbf{Training cost comparison.} We roughly estimate the training cost of each model. $\ddagger$ indicates concurrent works.} 
    \label{fig:cost}
    \vspace{-2pt}
\end{figure}

\section{Discussion}

\subsection{Analysis}

\noindent
\textbf{Training cost.} We show the estimated training cost comparison in Fig.~\ref{fig:cost}, where 1 H100 hour counts as 3 A100 hours. Since $\pi$3 did not report their overall training time, we roughly estimate their training cost based the number of training epochs and GPUs used for training. The training of our model requires around 80 H100 hours. Compared to $\pi$3~\cite{wang2025pi3} and MapAnything~\cite{wang2025pi3}, our model requires significantly less add-on cost on top of VGGT~\cite{wang2025vggt}.

\noindent
\textbf{Ablation studies.} Tab.~\ref{tab:ablation} compares our compact 3D backend with a 2D variant using alternating attention. Our 3D backend consistently outperforms the 2D one, highlighting the advantage of maintaining a sparse yet compact 3D scene representation for explicit geometric reasoning.

\begin{table}[t!]
\centering
\setlength{\tabcolsep}{8pt}
\resizebox{\columnwidth}{!}{
\begin{tabular}{l ccc ccc ccc}
\toprule
\multirow{2}{*}{\textbf{Method}}
& \multicolumn{3}{c}{\textbf{\ethdshort{}}}
& \multicolumn{3}{c}{\textbf{\dtushort{}}}
& \multicolumn{3}{c}{\textbf{7-Scenes}}
\\ \cmidrule(lr){2-4} \cmidrule(lr){5-7} \cmidrule(lr){8-10}
& $\absrel\!\downarrow$ & Acc$\!\downarrow$ & Cp$\!\downarrow$
& $\absrel\!\downarrow$ & Acc$\!\downarrow$ & Cp$\!\downarrow$
& $\absrel\!\downarrow$ & Acc$\!\downarrow$ & Cp $\!\downarrow$
\\ \midrule

w/o backend &
6.02& 12.81& 11.89 &
0.83& \bf 0.22 & 0.08 &
5.51 & 2.32 & 3.51 \\

w 2D backend &
5.32& 11.78& 12.78 &
0.92& 0.23 & 0.19 &
5.15 & 1.92 & 3.10 \\

Full &
\bf 4.64& \bf 9.98& \bf 9.69 &
\bf 0.81& 0.22 & \bf 0.08 &
\bf 4.74 & \bf 1.74 & \bf 2.84 \\

\bottomrule
\end{tabular}}
\vspace{-5pt}
\caption{\textbf{Ablation study.} We ablate on our 3D backend versus 2D backend implemented by alternating attention. }
\label{tab:ablation}
\vspace{-2pt}
\end{table}

\subsection{Conclusion}

We presented \papername{}, a feed-forward model that integrates a sparse yet compact volumetric scene representation as its backend.  Our paper shows this spatial compactness leads to state-of-the-art performance across diverse 3D vision tasks—including monocular/multi-view metric-scale depth/3D reconstruction, uncalibrated visual odometry and structure-from-motion. Notably, all these results are achieved without test time optimization or task-specific fine-tuning. This shows a step towards a scalable, unified, and generalizable feed-forward 3D perception system.
\clearpage
\maketitlesupplementary

\section{Limitations}

\subsection{Scene Representations}

\noindent
\textbf{Local scene representation.} Our backend representation is a local scene representation that cannot perform joint reasoning across multiple chunks predicted by the front-end. Extending it to a global representation  can be a promising direction as it would enable long-term consistency across chunks. However, training such a global backend would also require substantially more computational resources.

\noindent
\textbf{Computation complexity.} Our model shares the inherent limitation of type-(c) methods regarding the quadratic computational complexity with respect to the number of input images. However, a key advantage of our sparse yet compact 3D representation is that its complexity depends on the amount of 3D content rather than the number of views. This is, in fact, one of our motivations to introduce a compact 3D scene representation as a backend. Due to the computational constraint, we are not able to investigate on how to re-design the entire pointmap-based foundation models. However, our work serves as a proof-of-concept demonstrating the effectiveness of the spatial compactness. We believe this opens a promising direction towards making pointmap-based foundation models themselves more scalable to longer sequences. One possible direction is to reduce the usage of global attention and leverage the compact 3D scene representation as an alternative to reduce the computational complexity for long sequences. 

\noindent
\textbf{Dynamic scenes.} Since our model has not been trained for dynamic scenarios, it heavily relies on statics cues for dynamic environment. Thus, it might fail when the target scene is dominated by dynamic objects. This can be solved by including dynamic scenes as training datasets.

\subsection{Visual Odometry}

\noindent
\textbf{Reliance on dense reconstruction prior.} Our visual odometry relies on the prior that the model predicts geometry in reference (first) frame’s coordinate system, up to an unknown scale factor. When this assumption breaks, typically in scenes with diverse depth ranges and complex thin structures, the scale alignment becomes unreliable, often leading to tracking failure. Similarly, scenes dominated by distant content can increase prediction variance, resulting in poor scale alignment and trajectory accuracy. In this case, type-(b) method might benefit from its persistent memory and a generally longer perception window.

\noindent
\textbf{Drift/Kidnapping issue.} Our system is a visual odometry without explicit loop closure or relocalization. Consequently, tracking might drift in large-scale environments or fail under long-term kidnapping scenarios. In this case, an optimization-based module could help for loop closure or relocalization. Noteably, our backward search strategy in keyframe selection could also mitigate this issue. Once loop is closed, as long as the accumulative pose error is less than the backward search pose distance, our model can still leverage the earlier frame to bring the system back on track. This is useful for an online system, as it is not necessary to fix the earlier trajectory in that case.

\subsection{Structure from Motion}

\noindent
\textbf{Reliance on dense reconstruction prior.} Our SfM shares the same limitations as in VO as they rely on the same prior.

\noindent
\textbf{Initialization.} Our initialization is purely based on the feature-based image clustering and the confidence of the local chunk. We notice that our confidence usually prefers indoor regions. In some photo-tourism scenarios, our model might initialize with indoor images that do not guarantee overlap with the outdoor regions that are of primary interest. One possible solution is to use explicit feature matching to construct the view graph at cost of extra complexity.

\noindent
\textbf{Mapping window selection.} Our mapping window selection for global mapping is based on pose distance. This is usually sufficient for visual odometry due to the local smoothness of the trajectory. However, for a large-scale structure from motion system. The same geometry is likely to be observed from diverse views. Selecting mapping frames using pose distance can therefore omit overlapping views with wide baselines. In that case, a geometry-based view selection might help at the cost of extra complexity.

\noindent
\textbf{Non-overlap reconstruction.} The ability to reconstruct non-overlapping views is useful for small-scale problems where overlap genuinely does not exist. However, this could become problematic at larger scales: image clustering may group visually similar images from entirely different locations into the same cluster despite having no geometric overlap. In such cases, the model may hallucinate geometry for these non-overlapping images and give them moderate confidence, preventing them from being pruned and leading to irreversible reconstruction errors. One possible remedy is to filter out false clusters (e.g., Doppelgangers \cite{cai2023doppelgangers}).

\noindent
\textbf{Pose consistency.} The final camera poses are weighted average feed-forward poses. Compared to COLMAP~\cite{schonberger2016colmap} poses obtained via BA, our poses may not strictly follow the geometry constraint. For downstream tasks like novel view synthesis, which requires strict pose coherence, our poses, even metrically better in some cases, might still result in lower PSNR. This is the inherent limitation of feed-forward poses and can be addressed via BA as post-processing.

\section{Additional Details}

\subsection{Metric-scale Head}

To predict metric scale factor, we first map encoder feature into one feature vector via a three-layer MLP. We then add this feature with depth feature from depth DPT~\cite{ranftl2021dpt} branch of VGGT~\cite{wang2025vggt}, and map this new feature into a metric scale factor via a two-layer convolution. We supervise this prediction using an $\gL_1$ loss. In cases where the selected pixels contain missing depth values, which frequently occurs in the Waymo~\cite{sun2020waymo} dataset, we use a ROE solver to estimate the relative scale difference. This allows us to recover a consistent ground-truth log-depth value for supervision even when raw depth is unavailable.

\subsection{Training Datasets}
We use ScanNet~\cite{dai2017scannet}, ScannetPP~\cite{yeshwanthliu2023scannetpp}, WildRGBD~\cite{xia2024wildrgbd}, Mapfree~\cite{arnold2022mapfree}, Aria~\cite{pan2023aria}, Waymo~\cite{sun2020waymo}, Virtual Kitti2~\cite{cabon2020vkitti2}, GTASfM~\cite{wang2020gtasfm}, MVS-Synth~\cite{huang2018mvssynth}, OminiObject3d~\cite{wu2023omniobject3d}, and Hypersim~\cite{roberts2021hypersim} to train our model. Due to the constraint of the data storage, we only store a subset of each dataset. Although we only sample 2k samples in total for each epoch, we find data diversity is significantly more important than the data amount for training pointmap-based foundation model. We exclude Co3D~\cite{reizenstein2021common} as we observe that VGGT might overfit on certain patterns on Co3D.

\begin{figure}[t]
    \centering
    \includegraphics[width=\columnwidth]{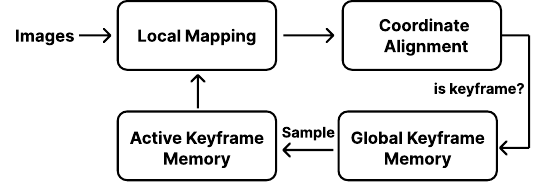}
    \vspace{-5pt}
    \caption{\textbf{Overview of \papername{} (VO).} Input frames are mapped with the keyframes in the active keyframe memory to predict geometry and camera poses. After coordinate alignment, we select new keyframes and update the global keyframe memory; poses and geometry for non-keyframes are also stored. If the active keyframe memory is not full, the new keyframe is appended; otherwise, we refresh the active keyframe memory by resampling a new set of keyframes from the global keyframe memory.}
    \label{fig:sup_vo_pipe}
    \vspace{-2pt}
\end{figure}

\subsection{Visual Odometry Pipeline}

Fig.~\ref{fig:sup_vo_pipe} illustrates our visual odometry pipeline. Input frames are mapped with keyframes stored in the active keyframe memory to predict camera poses and geometry. The coordinate alignment is done via 1) transforming the active keyframe map from global to local coordinates 2) estimating the relative scale of the corresponding keyframe geometry 3) transforming the local map to global coordinates via the weighted average of relative poses of each corresponding keyframe. We then select new keyframes from the newly mapped frames, and update the global keyframe memory. If the number of keyframes in active keyframe memory has not reached its capacity, we append the new keyframe; otherwise, we refresh the entire active keyframe memory by resampling from the global keyframe memory.

\begin{figure*}[t]
    \centering
    \includegraphics[width=\textwidth]{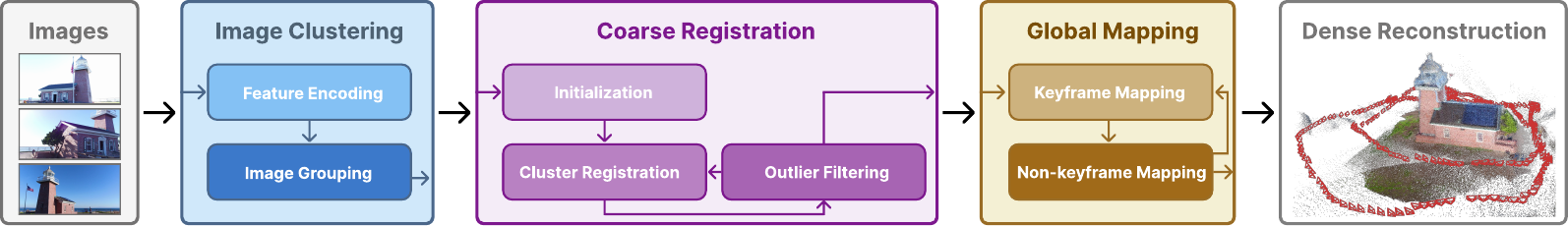}
     \vspace{-15pt}
    \caption{\textbf{Overview of \papername{} (SfM).} Our SfM pipeline contains 1) Image clustering that groups images into small clusters, 2) Coarse registration that constructs an initial coarse reconstruction, and 3) Global mapping that performs keyframe and non-keyframe refinement.} 
    \label{fig:sup_sfm_pipeline}
    \vspace{-10pt}
\end{figure*}

\subsection{Structure from Motion Pipeline}

We show the pipeline of our SfM as in Fig.~\ref{fig:sup_sfm_pipeline}. Our SfM mainly consists of 3 stages: 1) Image clustering that groups images into small clusters, 2) Coarse registration that registers each cluster incrementally, and 3) Global mapping that refines keyframe and non-keyframes via mapping.

\subsection{Visual Odometry Runtime Analysis}

We evaluate the runtime of our visual odometry on the TUM dataset~\cite{sturm2012tumrgbd}. Note that we exclude the data-loading overhead. Our method runs at an average of 4.2 FPS, with a best case of 6.0 FPS and a worst case of 3.4 FPS on an NVIDIA RTX 4090 GPU with input resolution of (392, 518). The variation in speed is primarily caused by the fluctuating number of active keyframes and the scene difficulty (The backend can be skipped if the front-end confidence is sufficiently high). Importantly, since we cap the maximum number of active keyframes at 10, \textbf{the computational complexity does not grow with respect to number of frames}, unlike type-(b) methods that do not have memory pruning.

\begin{figure}[t]
    \centering
    \includegraphics[width=\columnwidth]{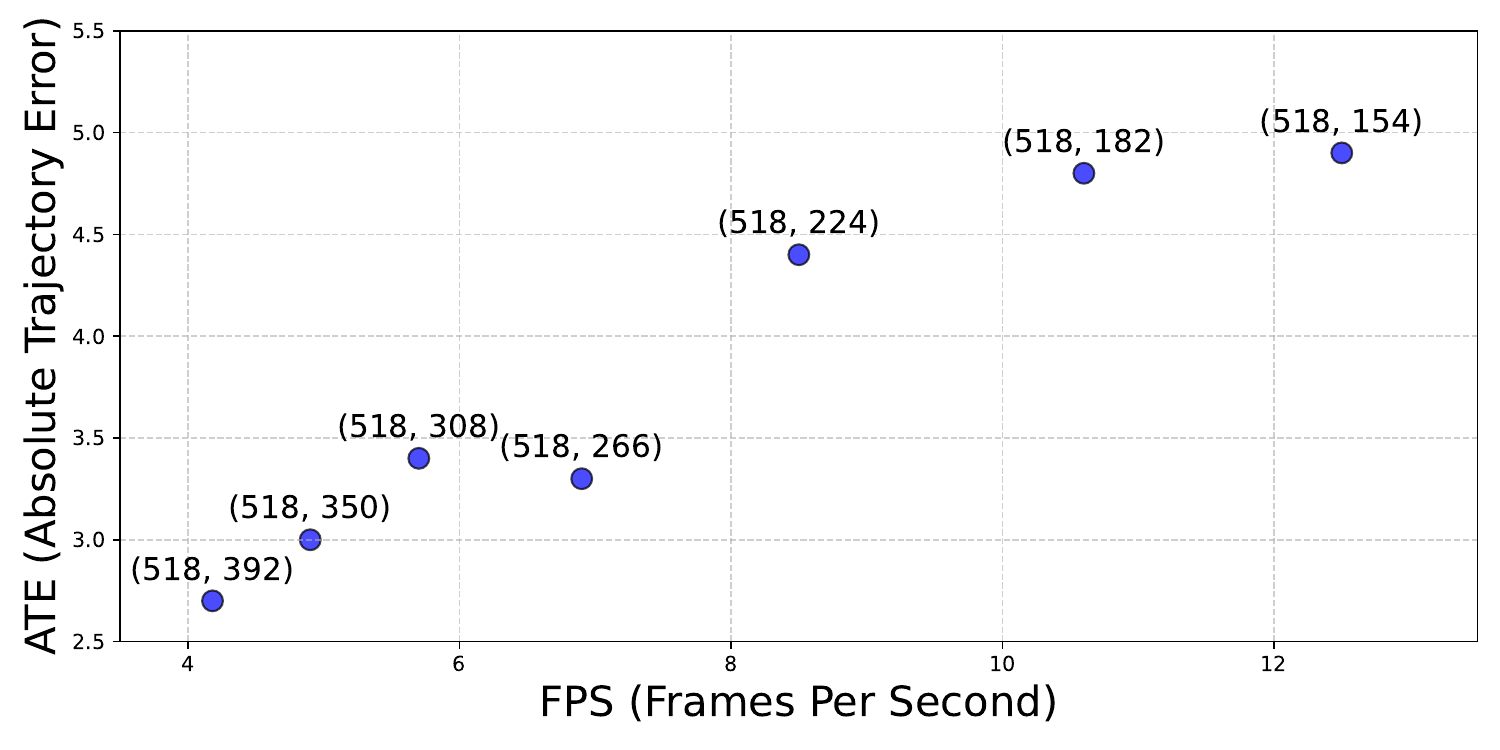}
    \vspace{-18pt}
    \caption{\textbf{VO runtime with respect to resolution.} By reducing input resolution, our method can achieve over 10FPS inference while being more accurate compared to other uncalibrated VO.} 
    \label{fig:sup_vo_time}
    \vspace{-2pt}
\end{figure}

Depending on the computation budget, one might reduce the input resolution or number of active keyframes for better runtime at the cost of accuracy (See Fig.~\ref{fig:sup_vo_time}).

\subsection{Imperfect Ground-truth on 7 Scenes}

We find that the pseudo ground truth in the original 7-Scenes~\cite{shotton2013scorf} dataset often contains noticeable drift in certain sequences. This is because those poses are obtained via ICP-based KinectFusion~\cite{newcombe2011kinectfusion}. ICP is prone to have rotation drift in scenes with near-spherical geometry and translation drift in scenes with many flat surfaces. 7 scenes dataset is the latter case. For instance, in Pumpkin/seq01, we find a substantial drift toward the end of the sequence. As a result, it is fundamentally impossible to achieve accurate tracking when evaluated against this pseudo GT unless the model happens to drift in exactly the same way. This makes the original pseudo GT unsuitable for fair comparison.

To address this, we adopt COLMAP GT~\cite{brachmann2021newgt7scene} that is obtained via global optimization across all sequences of each scene. To ensure its reliability, we follow ACE-Zero~\cite{brachmann2024scene} and evaluate novel view synthesis using NeRFstudio~\cite{nerfstudio}. As shown in Tab.~\ref{tab:slam_7scenes}, the PSNR obtained from rendering with the new GT is consistently higher, confirming its accuracy. We therefore use it as the new ground-truth for evaluating 3D reconstruction and visual odometry.

\subsection{Baselines and Evaluation Datasets}

\textbf{Monocular depth estimation.} We compare against Ominidata~\cite{eftekhar2021omnidata}, Depth Anything v2~\cite{yang2024depthanythingv2}, Marigold~\cite{ke2023marigold}, Diffusion-E2E~\cite{garcia2025diffusione2e}, and MoGe~\cite{wang2025moge}, which are explicitly trained for monocular depth estimation. In contrast, VGGT~\cite{wang2025vggt} and our model are trained with a multi-view objective, and evaluated on monocular depth in a zero-shot manner. We use NYUv2~\cite{silberman2012nyuv2}, KITTI~\cite{geiger2012kitti}, ETH3D~\cite{schops2017eth3d}, ScanNet~\cite{dai2017scannet}, and DIODE~\cite{vasiljevic2019diode} for evaluation following Marigold~\cite{ke2023marigold}. Note that for DIODE, there is a known issue about floaters in ground-truth geometry. Existing works like depth anything v2~\cite{yang2024depthanythingv2} and MoGe~\cite{wang2025moge} evaluate on DIODE with a pre-processing script that excludes those floaters. Since these scripts are not publicly available, we evaluate on the original noisy ground-truth following Marigold and Diffusion-E2E. 

\noindent
\textbf{Multi-view depth estimation.} We compare with 5 different categories of methods here: a) classic approach: COLMAP~\cite{schonberger2016colmap}, which is an integrated solution of SfM and MVS based on optimization, b) monocular depth estimation methods that estimate depth from a single view: Depth Pro~\cite{bochkovskiy2025depthpro}, Metric3D~\cite{yin2023metric3d}, UniDepthV2~\cite{piccinelli2025unidepthv2}, and Depth Anything v2~\cite{yang2024depthanythingv2}, c) Multi-view depth estimation methods that requires known calibration, camera poses, and per-image range: MVSNet~\cite{yao2018mvsnet}, Vis-MVSNet~\cite{zhang2023vismvsnet}, PatchmatchNet~\cite{wang2021patchmatchnet}, and MVSFormer++~\cite{cao2024mvsformerplus}. d) Multi-view depth estimation methods without the need for per-image range: Fast-MVSNet~\cite{yu2020fastmvsnet}, Robust MVD baseline~\cite{schroppel2022rmvd}, and MVSA~\cite{izquierdo2025mvsanywhere}. and e) Depth estimation from raw images without any prior information required: DeMoN~\cite{ummenhofer2017demon}, DUSt3R~\cite{wang2024dust3r}, Spann3R~\cite{wang2025spann3r}, Pow3R~\cite{jang2025pow3r}, MUSt3R~\cite{cabon2025must3r}, VGGT~\cite{wang2025vggt}, as well as concurrent works $\pi$3~\cite{wang2025pi3}, and MapAnything~\cite{keetha2025mapanything}. We evaluate those methods on RMVDB~\cite{schroppel2022rmvd}, using KITTI~\cite{geiger2012kitti}, ETH3D~\cite{schops2017eth3d}, ScanNet~\cite{dai2017scannet}, DTU~\cite{aanaes2016dtu}, and Tanks \& Temples~\cite{Knapitsch2017tankandtemple} datasets.

\begin{figure}[t]
    \centering
    \includegraphics[width=0.99\columnwidth]{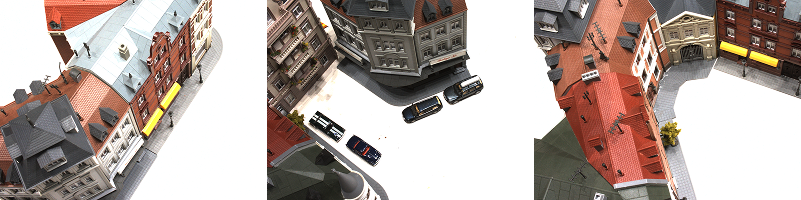}
     \vspace{-5pt}
    \caption{\textbf{Examples} of buildings in DTU~\cite{aanaes2016dtu} dataset.} 
    \label{fig:sup_dtu_building}
    \vspace{-10pt}
\end{figure}

\noindent
\textbf{Metric-scale estimation.} We compare with existing pointmap-based foundation models which can recover metric-scale factors: MASt3R~\cite{leroy2024mast3r}, 
Spann3R~\cite{wang2025spann3r}, MUSt3R~\cite{cabon2025must3r}, CUT3R~\cite{wang2025cut3r}, and concurrent work MapAnything~\cite{keetha2025mapanything} on RMVDB~\cite{schroppel2022rmvd}. Note that DTU datasets contains many buildings placed on a pure white table as in Fig.~\ref{fig:sup_dtu_building}. Due to the lack of background environment for scale reasoning, existing models would usually predict those buildings as real buildings, resulting in high absolute relative errors. In that case, inlier ratio could be a good measure of the metri-scale prediction.

\noindent
\textbf{3D reconstruction.} We compare with existing representative pointmap-based foundation models: Spann3R~\cite{wang2025spann3r}, MUSt3R~\cite{cabon2025must3r}, CUT3R~\cite{wang2025cut3r}, VGGT~\cite{wang2025vggt}, as well as concurrent works $\pi$3~\cite{wang2025pi3}, and MapAnything~\cite{keetha2025mapanything} on ETH3D~\cite{schops2017eth3d}, DTU~\cite{aanaes2016dtu}, and 7 scenes~\cite{shotton2013scorf}. We use image tuples from RMVDB~\cite{schroppel2022rmvd} for ETH3D and DTU. For 7 scenes, we use Spann3R~\cite{wang2025spann3r} split with improved GT.

\noindent
\textbf{Video depth estimation.} We compare with pointmap-based models: Spann3R~\cite{wang2025spann3r}, CUT3R~\cite{wang2025cut3r}, VGGT~\cite{wang2025vggt}, and $\pi$3~\cite{wang2025pi3} on dynamic video depth estimation task on Sintel~\cite{butler2012sintel}, Bonn~\cite{palazzolo2019bonn}, and Kitti~\cite{geiger2012kitti} datasets. Among those methods, CUT3R~\cite{wang2025cut3r} and $\pi$3~\cite{wang2025pi3} are trained on dynamic datasets while our model  is not specifically trained on dynamic data.

\noindent
\textbf{VO and SLAM.} For visual odometry, we compare with 1) Sparse VO: ORB-SLAM3~\cite{mur2015orb}, DSO~\cite{engel2017dso}, and DPVO~\cite{lipson2024dpvslam}, 2) Dense VO: TANDEM~\cite{koestler2022tandem}, MonoGS~\cite{matsuki2024monogs}, DeepFactors~\cite{czarnowski2020deepfactors}, DepthCov~\cite{dexheimer2023depthcov}, DROID-VO~\cite{teed2021droidslam}, COMO~\cite{dexheimer2024como}, and GLORIE-VO~\cite{zhang2024glorieslam}, and 3) Uncalibrated VO: Spann3R~\cite{wang2025spann3r} and MUSt3R~\cite{cabon2025must3r}. In addition to visual odometry baselines, we also consider SLAM baselines with global bundle adjustment and loop closure: 1) SLAM with calibration: ORB-SLAM3~\cite{mur2015orb}, DeepV2D~\cite{teed2020deepv2d}, DeepFactors~\cite{czarnowski2020deepfactors}, DPV-SLAM~\cite{lipson2024dpvslam}, GO-SLAM~\cite{zhang2023goslam}, DROID-SLAM~\cite{teed2021droidslam}, MASt3R-SLAM~\cite{murai2025mast3rslam} and 2) SLAM without calibration: DROID-SLAM~\cite{teed2021droidslam}, MASt3R-SLAM~\cite{murai2025mast3rslam}, VGGT-SLAM~\cite{wang2025vggt}. We compare with those methods using common SLAM benchmarks, including TUM~\cite{sturm2012tumrgbd}, ETH-SLAM~\cite{schops2019badslam}, and 7scenes~\cite{shotton2013scorf}. We also compare with MegaSaM~\cite{li2025megasam}, a structure-and-motion methods specifically designed for dynamic environment, on TUM Dynamic~\cite{sturm2012tumrgbd} dataset to test the generalization on dynamic scenes.

\noindent
\textbf{Structure from Motion.} We compare with optimization-based SfM methods: COLMAP~\cite{schonberger2016colmap}, ACE-Zero~\cite{brachmann2024scene}, FlowMAP~\cite{smith2024flowmap}, VGGSfM~\cite{wang2025vggt}, DF-SfM~\cite{he2024dfsfm}, and MASt3R-SfM~\cite{duisterhof2025mast3rsfm} on ETH3D~\cite{schops2017eth3d} and Tanks\&Temples~\cite{Knapitsch2017tankandtemple} dataset. ETH3D~\cite{schops2017eth3d} contains unordered image collection while Tanks\&Temples~\cite{Knapitsch2017tankandtemple} contains images from video.

\begin{table}[t!]
\centering
\setlength{\tabcolsep}{3pt}
\resizebox{1.0\columnwidth}{!}{
\begin{tabular}{l cc cc cc cc cc cc}

\toprule
    \textbf{Method}
    & \multicolumn{2}{c}{\textbf{\kittishort{}}}
    & \multicolumn{2}{c}{\textbf{\scannetshort{}}}
    & \multicolumn{2}{c}{\textbf{\ethdshort{}}}
    & \multicolumn{2}{c}{\textbf{\dtushort{}}}
    & \multicolumn{2}{c}{\textbf{\tanksandtemplesshort{}}}
    & \multicolumn{2}{c}{\textbf{Avg}}
    \\
    \cmidrule(lr){2-3} \cmidrule(lr){4-5} \cmidrule(lr){6-7} \cmidrule(lr){8-9} \cmidrule(lr){10-11} \cmidrule(lr){12-13}

    & $\absrel\downarrow$ & $\delta\uparrow$
    & $\absrel\downarrow$ & $\delta\uparrow$
    & $\absrel\downarrow$ & $\delta\uparrow$
    & $\absrel\downarrow$ & $\delta\uparrow$
    & $\absrel\downarrow$ & $\delta\uparrow$
    & $\absrel\downarrow$ & $\delta\uparrow$
    \\
    \midrule

    w/o backend &  4.5 & 60.4 & (2.3) & (80.8) & 1.8 & 85.3 & 1.0 & 94.8 & 2.0 & 83.9 & 2.3 & 80.6 \\

    w/ 2d backend &  2.9 & 73.9 & (2.0) & (84.2) & 1.4 & 90.2 & 1.1 & 94.4 & 1.8 & 88.4 & 1.8 & 86.2 \\

    w/o scale align &  3.0 & 72.0 & (2.0) & (84.8) & 1.5 & 89.7 & \bf 0.9 & \bf 95.3 & 1.9 & 89.0 & 1.9  & 86.2 \\

    w/o zero conv &  26.9 & 7.8& (17.5) & (15.5) & 19.8 & 17.8 & 7.3 & 34.6 & 16.2 & 23.9 &  17.5 & 19.9  \\

    Full & \bf 2.8 & \bf 74.4 & (\bf1.9) & (\bf85.8) & \bf1.4 & \bf90.9 & 0.9 & 95.1 & \bf1.7 & \bf 90.2 & \bf1.7 & \bf87.3 \\

\bottomrule
\end{tabular}}
\caption{\textbf{Ablation studies}. We ablate design choices of backend, scale alignment for supervision, and zero convolution on RMVDB~\cite{schroppel2022rmvd}.}
\label{tab:sup_ab}
\end{table}

\definecolor{bgcolor}{RGB}{190, 181, 190}
\definecolor{mylightgray}{RGB}{238,238,238} 
\colorlet{bgcolor}{mylightgray}

\begin{table*}[t!]
\centering
\resizebox{\textwidth}{!}{
\begin{tabular}{|l|c|c|c|c
|c >{\columncolor{bgcolor}} c
|c >{\columncolor{bgcolor}} c
|c >{\columncolor{bgcolor}} c
|c >{\columncolor{bgcolor}} c
|c >{\columncolor{bgcolor}} c
|c >{\columncolor{bgcolor}} c
|}

\hline
    \textbf{Approach}
    & \textbf{\scriptsize{GT}}
    & \textbf{\scriptsize{GT}}
    & \textbf{\scriptsize{GT}}
    & \textbf{Align}
    & \multicolumn{2}{c|}{\textbf{\kittishort{}}}
    & \multicolumn{2}{c|}{\textbf{\scannetshort{}}}
    & \multicolumn{2}{c|}{\textbf{\ethdshort{}}}
    & \multicolumn{2}{c|}{\textbf{\dtushort{}}}
    & \multicolumn{2}{c|}{\textbf{\tanksandtemplesshort{}}}
    & \multicolumn{2}{c|}{\textbf{Average}}
    \\

    & \textbf{\scriptsize{Poses}}
    & \textbf{\scriptsize{Range}}
    & \textbf{\scriptsize{Intrinsic}}
    &
    & $\absrel\downarrow$ & $\threshI\uparrow$
    & $\absrel\downarrow$ & $\threshI\uparrow$
    & $\absrel\downarrow$ & $\threshI\uparrow$
    & $\absrel\downarrow$ & $\threshI\uparrow$
    & $\absrel\downarrow$ & $\threshI\uparrow$
    & $\absrel\downarrow$ & $\threshI\uparrow$
    \\
    \hline
    \hline

    \multicolumn{17}{|l|}{\textbf{a) Classic approaches}}
    \\
    COLMAP~\cite{schonberger2016colmap}
	& \mn
	& \mn
	& \mn
	& \mn
	& 12.0
	& 58.2
	& 14.6
	& 34.2
	& 16.4
	& 55.1
	& \bf0.7
	& \bf96.5
	& 2.7
	& \bf95.0
	& 9.3
	& 67.8
	\\
    COLMAP Dense ~\cite{schonberger2016colmap}
	& \mn
	& \mn
	& \mn
	& \mn
	& 26.9
	& 52.7
	& 38.0
	& 22.5
	& 89.8
	& 23.2
	& 20.8
	& 69.3
	& 25.7
	& 76.4
	& 40.2
	& 48.8
	\\

    \hline
        \hline  
            \multicolumn{17}{|l|}{\textbf{b) Single-view depth}}
    \\

        Depth Pro~\cite{bochkovskiy2025depthpro}
        & \mn
        & \mn
        & \my
        & med
        &  6.1
        &  39.6
        &  (4.3)
        &  (58.4)
        &  6.1
        &  53.5
        &  5.6
        &  49.6
        &  5.6
        &  57.5
        &  5.6
        &  51.7
        \\

        Metric3D~\cite{hu2024metric3dv2}
        & \mn
        & \mn
        & \my
        & med
        &  5.1
        &  44.1
        &  2.4
        &  78.3
        &  4.4
        &  54.5
        &  10.1
        &  39.5
        &  6.2
        &  48.0
        &  5.6
        &  52.9
        \\
    
        UniDepthV2~\cite{piccinelli2025unidepthv2}
        & \mn
        & \mn
        & \my
        & med        
        &  4.0
        &  55.3
        &  (2.1)
        &  (82.6)
        &  3.7
        &  66.2
        &  3.2
        &  72.3
        &  3.6
        &  68.4
        &  3.3
        &  68.9
        \\

	DepthAnything V2~\cite{yang2024depthanythingv2}
	& \mn
	& \mn
	& \mn
	& lstsq $\dagger$
	& 6.6
	& 38.6
	& 4.0
	& 58.6
	& 4.7
	& 56.5
	& 2.6
	& 74.7
	& 4.5
	& 57.5
	& 4.8
	& 54.1
	\\ %

    \hline
    \hline

    \multicolumn{17}{|l|}{\textbf{c) Depth from frames and poses (w/ per-image range)}}
    \\

	MVSNet\normalsize{~\cite{yao2018mvsnet}}	& \my
	& \my
	& \my
	& \mn
	& 22.7
	& 36.1
	& 24.6
	& 20.4
	& 35.4
	& 31.4
	& (1.8)
	& (86.0)
	& 8.3
	& 73.0
	& 18.6
	& 49.4
	\\ %

	Vis-MVSNet~\cite{zhang2023vismvsnet}
	& \my
	& \my
	& \my
	& \mn
	& {9.5}
	& {55.4}
	& 8.9
	& 33.5
	& {10.8}
	& {43.3}
	& (1.8)
	& (87.4)
	& {4.1}
	& {87.2}
	& {7.0}
	& {61.4}
	\\ %

	PatchmatchNet~\cite{wang2021patchmatchnet}
	& \my
	& \my
	& \my
	& \mn
	& 10.8
	& 45.8
	& {8.5}
	& {35.3}
	& 19.1
	& 34.8
	& (2.1)
	& (82.8)
	& 4.8
	& 82.9
	& 9.1
	& 56.3
	\\ %

        MVSFormer++~\cite{cao2024mvsformerplus}
	& \my
	& \my
	& \my
	& \mn
        &  {4.4}
        &  {65.7}
        &  {7.9}
        &  {39.4}
        &  {7.8}
        &  {50.4}
        &  (0.9)
        &  (95.3)
        &  {3.2}
        &  {88.1}
        &  {4.8}
        &  {67.8}
        \\

    \hline
    \hline
    
    \multicolumn{17}{|l|}{\textbf{d) Depth from frames and poses (w/o per-image range)}}
    \\

	Fast-MVSNet~\cite{yu2020fastmvsnet}
	& \my
	& \mn
	& \my
	& \mn
	& 12.1
	& 37.4
	& 287.1
	& 9.4
	& 131.2
	& 9.6
	& (540.4)
	& (1.9)
	& 33.9
	& 47.2
	& 200.9
	& 21.1
	\\ %



	Robust MVD Baseline \cite{schroppel2022rmvd}
	& \my
	& \mn
	& \my
	& \mn
	& {7.1}
	& 41.9
	& {7.4}
	& {38.4}
	& {9.0}
	& {42.6}
	& {2.7}
	& {82.0}
	& {5.0}
	& {75.1}
	& {6.3}
	& {56.0}
	\\ %

        MVSA~\cite{izquierdo2025mvsanywhere}%
	& \my
	& \mn
	& \my
	& \mn
        &  3.2
        &  68.8
        &  3.7
        &  62.9
        &  3.2
        &  68.0
        &  1.3
        &  95.0
        &  2.1
        &  \underline{90.5}
        &  2.7
        &  77.0
        \\

        \hline
        \hline

        \multicolumn{17}{|l|}{\textbf{e) Depth from frames (w/o poses)}}
    \\
    
    DeMoN~\cite{ummenhofer2017demon}
    & \mn
    & \mn
    & \my
    & $\Vert \vect t \Vert$
    & 15.5
    & 15.2
    & 12.0
    & 21.0
    & 17.4
    & 15.4
    & 21.8
    & 16.6
    & 13.0
    & 23.2
    & 16.0
    & 18.3
    \\ %
    
    
    DUSt3R~\cite{wang2024dust3r} & \mn & \mn & \mn & med & 5.4 & 49.5 & (3.1) & (71.8) & 3.0 & 76.0 & 3.9 & 68.6 & 3.3 & 75.1 & 3.7 & 68.2 \\
    Spann3R~\cite{wang2025spann3r} & \mn & \mn & \mn & med & 7.9 & 36.2 & (3.3) & (67.1) & 5.7 & 58.6 & 3.5 & 65.2 & 4.7 & 58.5 & 5.0 & 57.1 \\
    Pow3R~\cite{jang2025pow3r} & \mn & \mn & \mn & med & 5.7 & 45.7 & (3.2) & (68.8) & 3.0 & 74.7 & 3.0 & 74.3 & 3.3 & 76.6 & 3.6 & 68.0 \\
    MUSt3R~\cite{cabon2025must3r} & \mn & \mn & \mn & med & 4.5 & 55.0 & (4.0) & (59.8) & 2.5 & 80.3 & 4.6 & 55.4 & (2.6) & (80.4) & 3.7 & 66.2\\

    VGGT~\cite{wang2025vggt} & \mn & \mn & \mn & med & 4.5 & 59.6 & (2.3) & (80.8) & 1.8 & 86.3 & \underline{0.9} & \underline{95.6} & 2.4 & 84.1 & 2.4 & 81.3 \\

    $\pi$3$^\ddagger$~\cite{wang2025pi3} & \mn & \mn & \mn & med & 2.8 & 72.9 & (2.0) & (83.6) & \bf 1.3 & \bf 92.4 & 1.3 & 91.8 & 1.8 & 87.3 & 1.8 & 85.6 \\

    MapAnything$^\ddagger$~\cite{wang2025vggt} & \mn & \mn & \mn & med & 4.0 & 59.4 & 4.0 & 60.5 & 2.8 & 73.2 & 3.9 & 63.7 & 3.3 & 73.0 & 3.6 & 66.0 \\

    \papernamecolor{} & \mn & \mn & \mn & med & \bf 2.8 & \textbf{74.4} & (\bf1.9) & (\bf85.8) & 1.4 & 90.9 & 0.9 & 95.1 & \bf1.7 & 90.2 & \bf1.7 & \bf87.3 \\

    \hline
\end{tabular}}
\vspace{-5pt}
\caption{\textbf{Multi-view depth estimation.} Our method achieves state-of-the-art performance on RMVDB~\cite{schroppel2022rmvd}. $\ddagger$ means concurrent works, and (parentheses) indicate. We only report a subset of a)-e) methods, and please refer to supplementary material for the full table.}
\label{tab:sup_mv_depth}
\end{table*}
\begin{table*}[t]
    \centering
    \setlength{\tabcolsep}{1pt} 
\resizebox{\textwidth}{!}{
\begin{tabular}{l l l r@{~}r r@{~}r r@{~}r r@{~}r r@{~}r r@{~}r r@{~}r}
\toprule
         & \multirow{2}{*}{} & \multirow{2}{*}{Scenes} & \multicolumn{2}{c}{COLMAP~\cite{schonberger2016colmap}} & \multicolumn{2}{c}{ACE-Zero~\cite{brachmann2024scene}} & \multicolumn{2}{c}{FlowMap~\cite{smith2024flowmap}} & \multicolumn{2}{c}{VGGSfM~\cite{wang2025vggt}} & \multicolumn{2}{c}{DF-SfM~\cite{he2024dfsfm}} & \multicolumn{2}{c}{MASt3R-SfM~\cite{duisterhof2025mast3rsfm}} & \multicolumn{2}{c}{\papernamecolor{} (SfM)} \\
         \cmidrule(lr){4-5} \cmidrule(lr){6-7} \cmidrule(lr){8-9} \cmidrule(lr){10-11} \cmidrule(lr){12-13} \cmidrule(lr){14-15} \cmidrule(lr){16-17}
           & & &   \multicolumn{1}{c}{\footnotesize RRA@5} &   \multicolumn{1}{c}{\footnotesize RTA@5} & \multicolumn{1}{c}{\footnotesize RRA@5} & \multicolumn{1}{c}{\footnotesize RTA@5} & \multicolumn{1}{c}{\footnotesize RRA@5} & \multicolumn{1}{c}{\footnotesize RTA@5} & \multicolumn{1}{c}{\footnotesize RRA@5} &   \multicolumn{1}{c}{\footnotesize RTA@5} & \multicolumn{1}{c}{\footnotesize RRA@5} & \multicolumn{1}{c}{\footnotesize RTA@5} & \multicolumn{1}{c}{\footnotesize RRA@5} & \multicolumn{1}{c}{\footnotesize RTA@5} & \multicolumn{1}{c}{\footnotesize RRA@5} & \multicolumn{1}{c}{\footnotesize RTA@5} \\
\midrule
& \multirow[c]{7}{*}{\rotatebox{90}{Training}} & {\small Barn} & {GT} & {GT} & \cellthird{56.1} & {55.6} & {-} & {-} & {-} & {-} & \cellfirst{100.} & \cellfirst{99.8} & {52.6} & \cellthird{85.6} & \cellsecond{91.8} & \cellsecond{96.4} \\
 &  & {\small Caterpillar} & {GT} & {GT} & \cellsecond{87.3} & \cellsecond{95.6} & {-} & {-} & {-} & {-} & {-} & {-} & \cellthird{84.2} & \cellthird{92.3} & \cellfirst{100.0} & \cellfirst{98.6} \\
 &  & {\small Church} & {GT} & {GT} & \cellsecond{90.5} & \cellsecond{76.3} & {-} & {-} & {-} & {-} & {-} & {-} & \cellthird{11.6} & \cellthird{16.8} & \cellfirst{93.0} & \cellfirst{92.9} \\
 &  & {\small Courthouse} & {GT} & {GT} & \cellsecond{44.1} & \cellsecond{45.0} & {-} & {-} & {-} & {-} & {-} & {-} & \cellthird{8.8} & \cellthird{9.9} & \cellfirst{79.9} & \cellfirst{77.7} \\
 &  & {\small Ignatius} & {GT} & {GT} & \cellfirst{100.} & \cellfirst{99.9} & {62.5} & {70.0} & {-} & {-} & \cellfirst{100.} & \cellfirst{99.9} & {43.6} & {60.1} & \cellfirst{100.0} & \cellsecond{99.5} \\
 &  & {\small Meetingroom} & {GT} & {GT} & {39.3} & {38.5} & {26.3} & {39.8} & {-} & {-} & \cellthird{84.1} & \cellthird{89.0} & \cellsecond{92.6} & \cellsecond{89.9} & \cellfirst{100.0} & \cellfirst{96.3} \\
 &  & {\small Truck} & {GT} & {GT} & \cellfirst{100.} & \cellsecond{99.7} & {53.4} & {69.6} & {-} & {-} & \cellfirst{100.} & \cellfirst{99.8} & \cellfirst{100.} & \cellsecond{99.7} & \cellfirst{100.0} & \cellsecond{99.7} \\
\cmidrule{2-17} 
 & \multirow[c]{8}{*}{\rotatebox{90}{Intermediate}} & {\small Family} & {GT} & {GT} & \cellsecond{38.9} & \cellsecond{44.6} & {-} & {-} & {-} & {-} & {-} & {-} & \cellthird{22.3} & \cellthird{25.9} & \cellfirst{100.0} & \cellfirst{99.6} \\
 &  & {\small Francis} & {GT} & {GT} & \cellthird{57.4} & \cellthird{79.0} & \cellsecond{57.6} & {67.7} & {-} & {-} & \cellfirst{100.} & \cellfirst{99.7} & {17.0} & {41.0} & \cellfirst{100.0} & \cellsecond{92.1} \\
 &  & {\small Horse} & {GT} & {GT} & \cellsecond{68.2} & \cellsecond{81.8} & {-} & {-} & {-} & {-} & {-} & {-} & \cellthird{6.4} & \cellthird{6.3} & \cellfirst{100.0} & \cellfirst{98.8} \\
 &  & {\small Lighthouse} & {GT} & {GT} & {30.6} & {38.8} & {4.8} & {9.5} & {-} & {-} & \cellsecond{66.3} & \cellthird{66.0} & \cellthird{50.8} & \cellsecond{72.1} & \cellfirst{100.0} & \cellfirst{98.7} \\
 &  & {\small M60} & {GT} & {GT} & \cellfirst{100.} & \cellsecond{99.9} & {50.4} & {48.3} & {-} & {-} & \cellfirst{100.} & \cellthird{99.8} & \cellfirst{100.} & \cellfirst{100.} & \cellfirst{100.0} & {99.0} \\
 &  & {\small Panther} & {GT} & {GT} & \cellfirst{100.} & \cellsecond{99.5} & \cellfirst{100.} & {77.6} & {-} & {-} & \cellfirst{100.} & {99.1} & \cellfirst{100.} & \cellsecond{99.5} & \cellfirst{100.0} & \cellfirst{99.6} \\
 &  & {\small Playground} & {GT} & {GT} & {82.7} & {85.5} & {49.1} & {63.8} & {-} & {-} & \cellfirst{100.} & \cellfirst{99.9} & \cellsecond{99.3} & \cellsecond{99.3} & \cellfirst{100.0} & \cellthird{98.3} \\
 &  & {\small Train} & {GT} & {GT} & \cellsecond{62.6} & \cellsecond{62.5} & {18.4} & {29.2} & {-} & {-} & \cellthird{42.8} & \cellthird{41.8} & {10.6} & {15.8} & \cellfirst{89.4} & \cellfirst{88.8} \\
\cmidrule{2-17} 
 & \multirow[c]{6}{*}{\rotatebox{90}{Advanced}} & {\small Auditorium} & {GT} & {GT} & {1.6} & {1.1} & {1.3} & {1.4} & {-} & {-} & \cellsecond{1.7} & \cellsecond{1.7} & \cellsecond{1.7} & \cellthird{1.5} & \cellfirst{96.0} & \cellfirst{88.7} \\
 &  & {\small Ballroom} & {GT} & {GT} & \cellfirst{56.4} & \cellsecond{43.2} & {14.1} & {16.7} & {-} & {-} & \cellsecond{56.0} & \cellfirst{44.4} & \cellthird{43.8} & {29.6} & {35.2} & \cellthird{37.5} \\
 &  & {\small Courtroom} & {GT} & {GT} & \cellthird{62.5} & \cellthird{54.1} & {5.3} & {3.6} & {-} & {-} & \cellsecond{66.8} & \cellsecond{66.3} & \cellfirst{67.2} & \cellfirst{69.1} & {38.7} & {51.1} \\
 &  & {\small Museum} & {GT} & {GT} & \cellthird{13.5} & \cellthird{11.1} & {0.8} & {1.2} & {-} & {-} & \cellsecond{14.8} & \cellsecond{13.5} & {12.3} & {11.0} & \cellfirst{100.0} & \cellfirst{98.4} \\
 &  & {\small Palace} & {GT} & {GT} & {3.1} & {3.9} & {-} & {-} & {-} & {-} & \cellthird{25.6} & \cellthird{27.7} & \cellsecond{27.0} & \cellsecond{35.7} & \cellfirst{38.6} & \cellfirst{61.4} \\
 &  & {\small Temple} & {GT} & {GT} & {0.4} & {0.9} & {0.5} & {1.2} & {-} & {-} & \cellthird{55.5} & \cellthird{60.7} & \cellsecond{80.7} & \cellsecond{72.2} & \cellfirst{99.3} & \cellfirst{96.9} \\
\bottomrule
\end{tabular}}
\caption{%
    \textbf{Detailed per-scene SfM results on Tanks and Temples~\cite{Knapitsch2017tankandtemple}.} We use COLMAP GT~\cite{Riegler2020FVS} for evaluation as in MASt3R-SfM~\cite{duisterhof2025mast3rsfm}. The baseline results are obtained from MASt3R-SfM. (-) indicates failure.
    }\label{tab:sup_sfm_tnt}

\end{table*}

\begin{figure*}[t]
    \centering
    \includegraphics[width=0.95\textwidth]{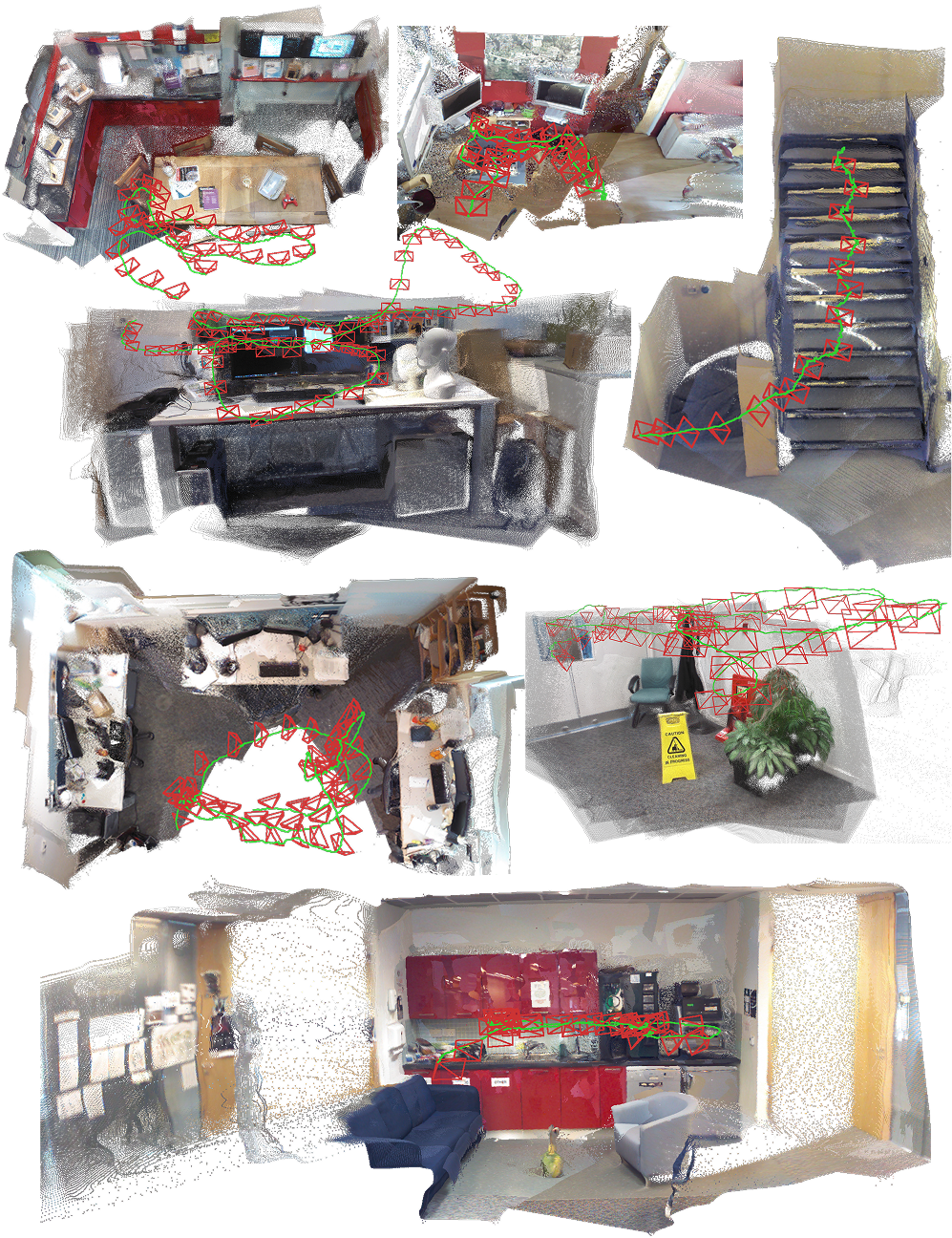}
    \caption{\textbf{Qualitative showcase} of VO on 7scenes~\cite{shotton2013scorf}. We visualize keyframe poses as red cameras and non-keyframe poses as green dots.} 
    \label{fig:sup_vo_seven}
    \vspace{-10pt}
\end{figure*}

\begin{figure*}[t]
    \centering
    \includegraphics[width=0.94\textwidth]{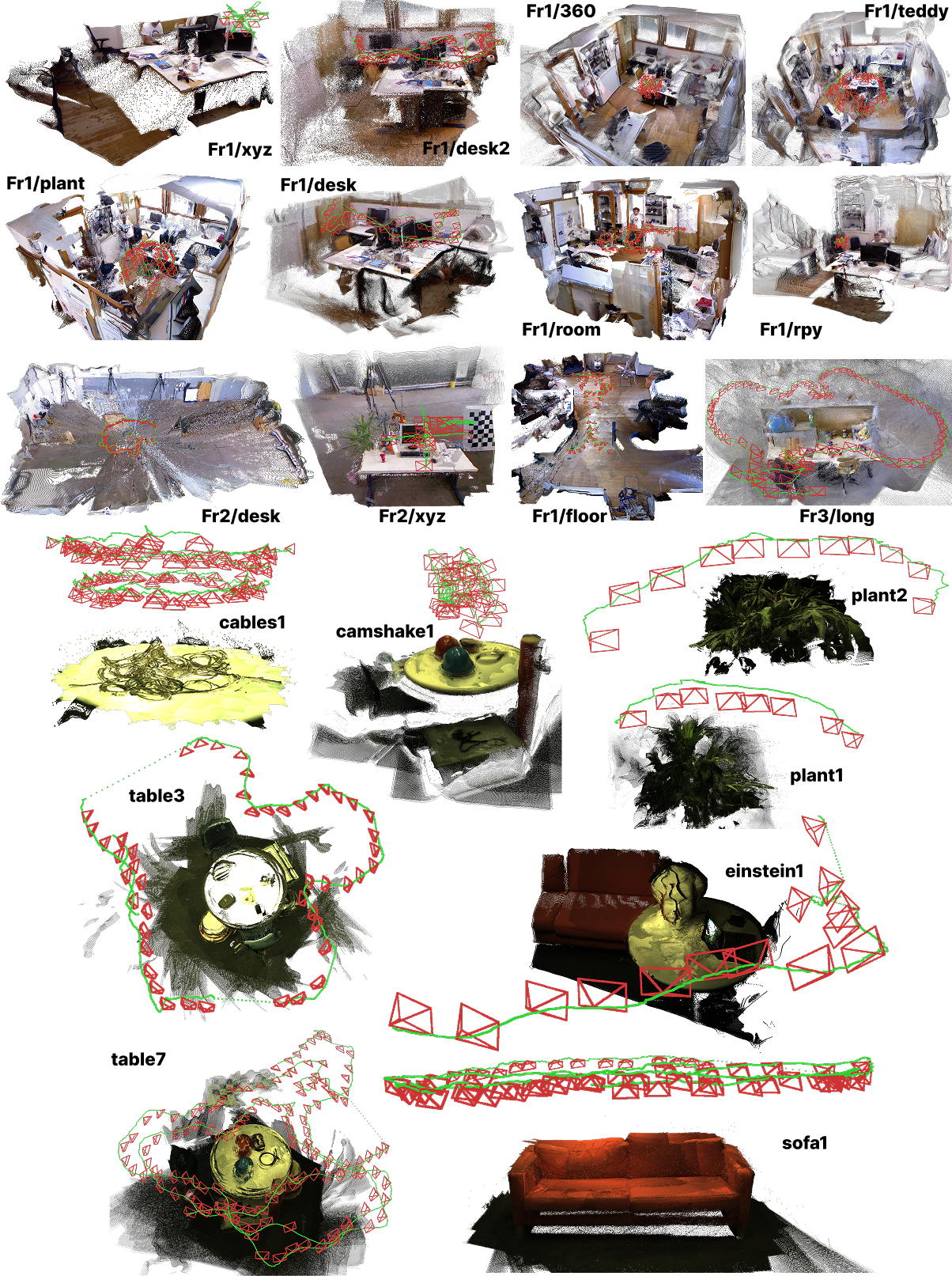}
    \caption{\textbf{Qualitative showcase} of VO on TUM~\cite{sturm2012tumrgbd} and ETH SLAM~\cite{schops2019badslam} datasets.} 
    \label{fig:sup_vo_tum}
    \vspace{-10pt}
\end{figure*}

\begin{figure*}[t]
    \centering
    \includegraphics[width=\textwidth]{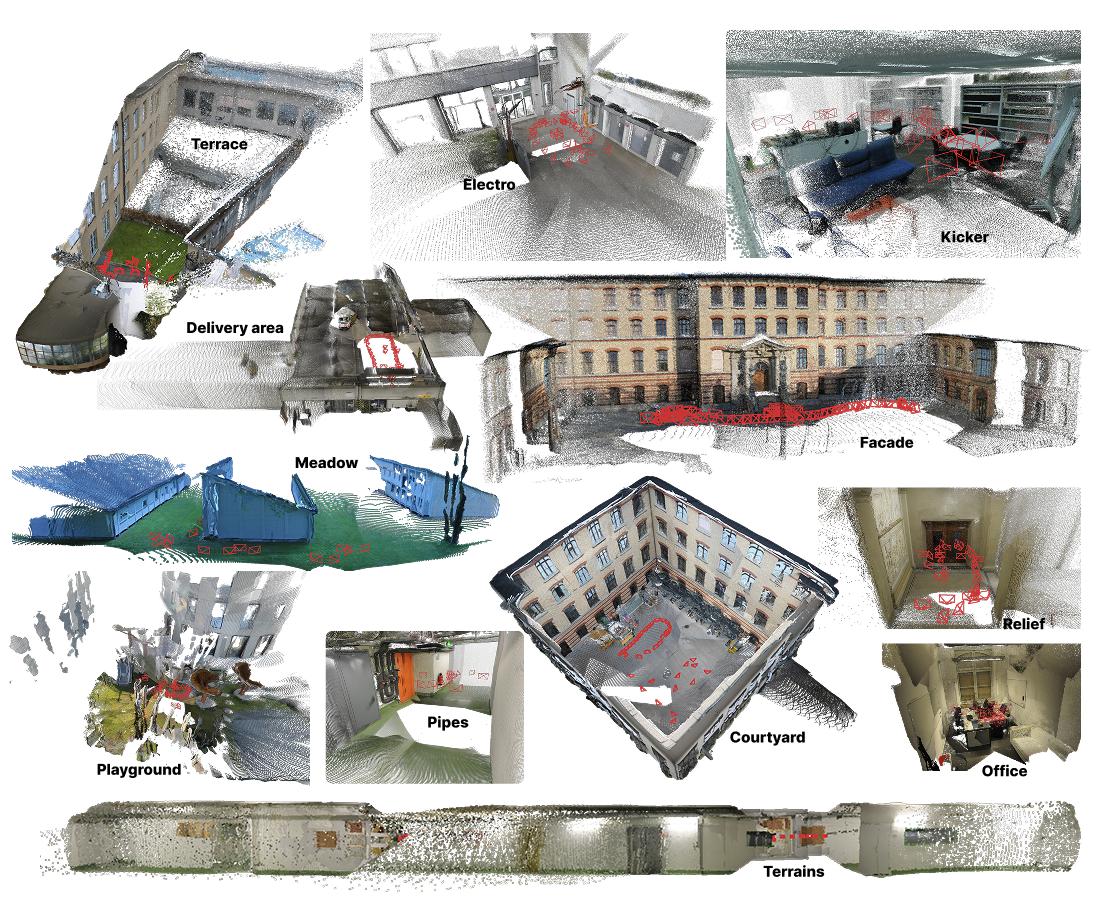}
     \vspace{-15pt}
    \caption{\textbf{Qualitative showcase} of structure from motion on ETH3D~\cite{schops2017eth3d}} 
    \label{fig:sup_sfm_eth}
    \vspace{-10pt}
\end{figure*}

\begin{figure*}[t]
    \centering
    \includegraphics[width=\textwidth]{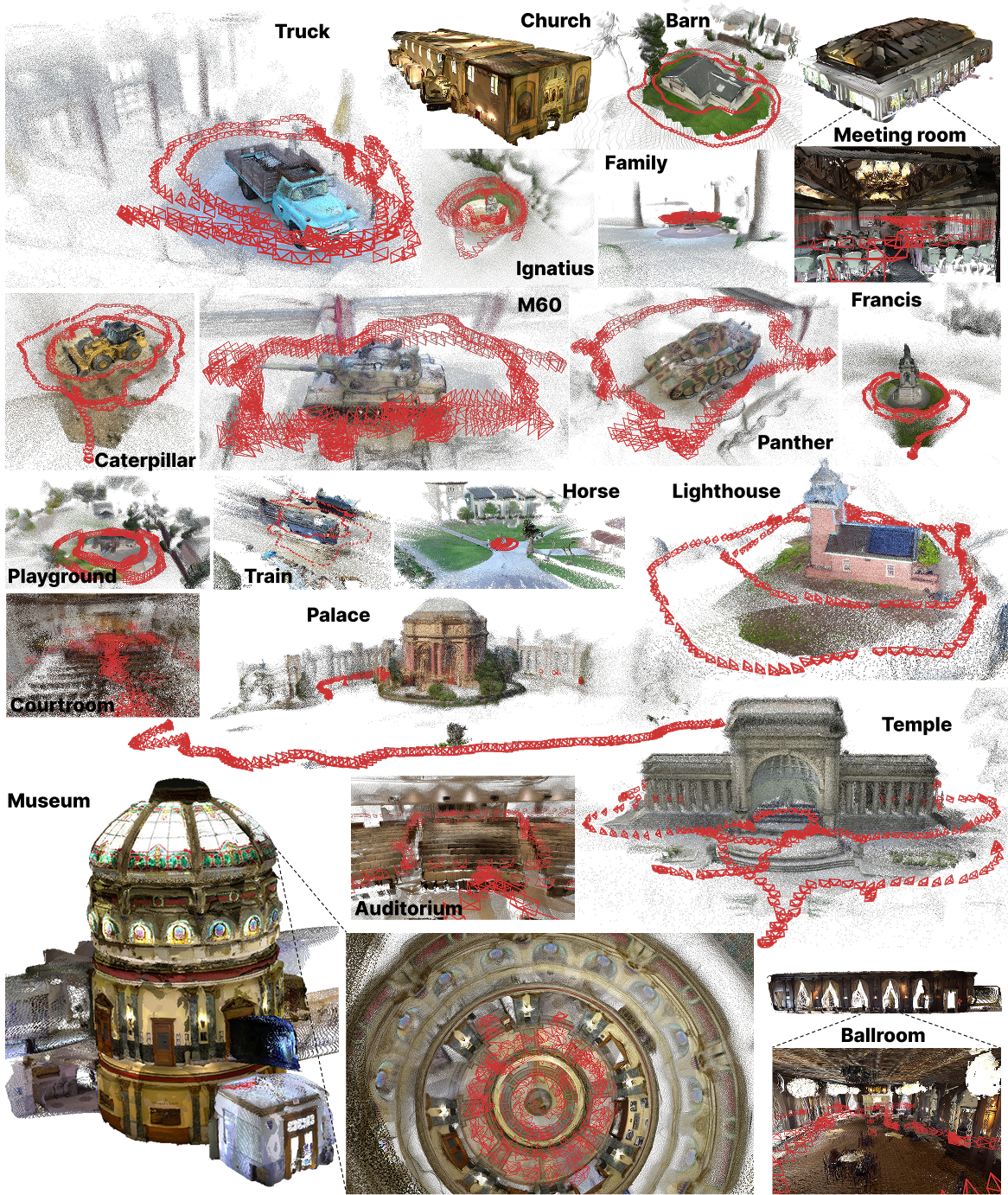}
    \caption{\textbf{Qualitative showcase} of structure from motion on Tanks and Temples~\cite{Knapitsch2017tankandtemple} dataset.} 
    \label{fig:sup_sfm_tnt}
\end{figure*}

\begin{figure*}[t]
    \centering
    \includegraphics[width=\textwidth]{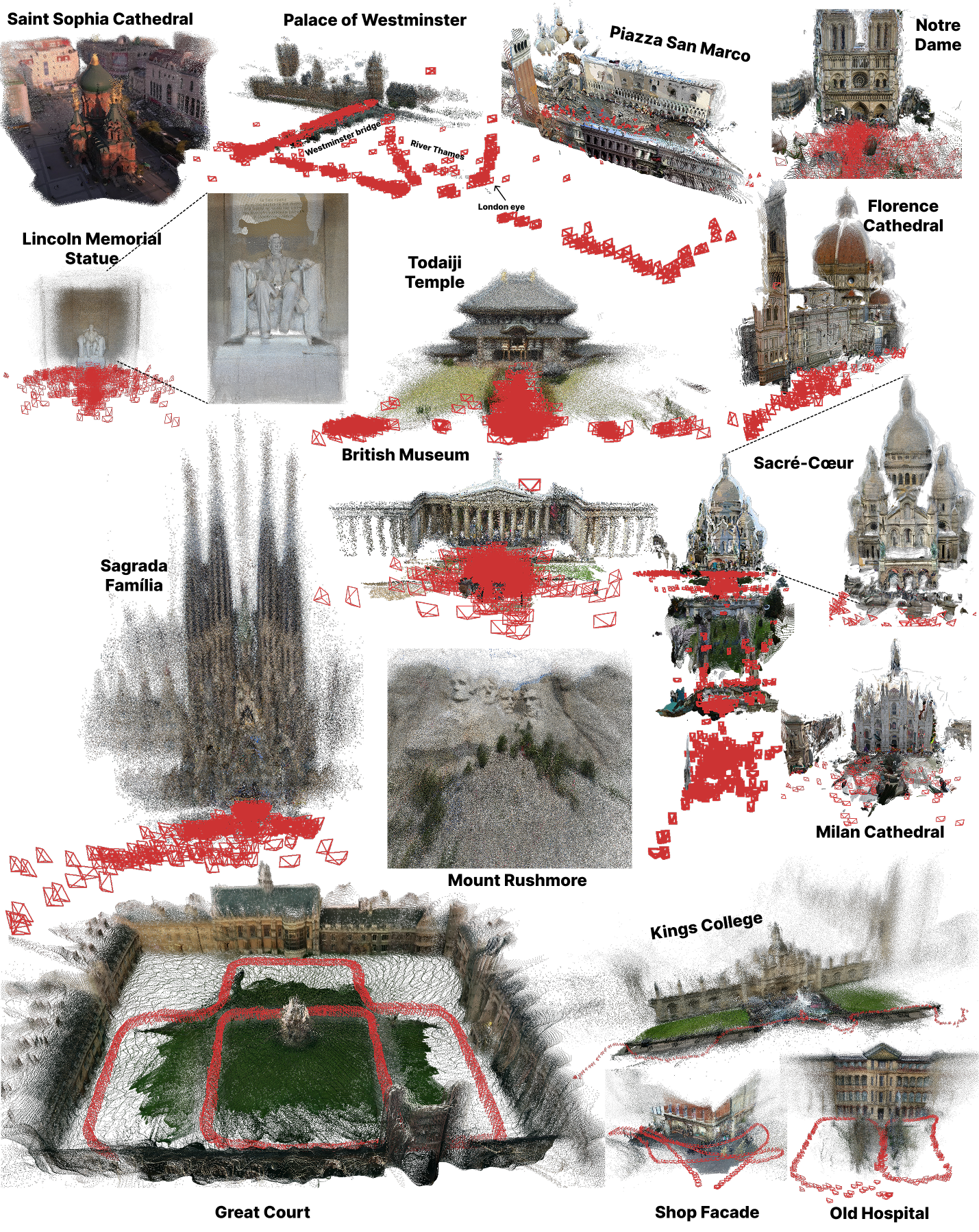}
     \vspace{-20pt}
    \caption{\textbf{Qualitative showcase} of structure from motion on IMC phototourism~\cite{jin2021imc}, Cambridge landmarks~\cite{kendall2015posenet}. The results are randomly downsampled to 3 million points for visualization purpose.} 
    \label{fig:sup_sfm_imc}
    \vspace{-10pt}
\end{figure*}

\section{Additional Quantitative Results}
\textbf{Ablation study.} We present additional ablation studies in Tab.~\ref{tab:sup_ab} to analyze the impact of key design choices, including the backend (2D vs. 3D), scale alignment for supervision, and zero convolution.  Notably, training without zero convolution fails to converge under our current computational budget and dataset size. As discussed in Sec.~\ref{sec:pre}, this is caused by catastrophic forgetting of the learned confidence: without zero convolution, the confidence function with randomly initialized weights of the backend will shift drastically, and lead to inconsistent learning objectives. In this case, convergence would likely require training resources and data comparable to those used for VGGT~\cite{wang2025vggt}. 

\noindent
\textbf{Multi-view depth estimation.} We show the full table of multi-view depth estimation results in Tab.~\ref{tab:sup_mv_depth} with detailed input modality and the alignment.

\noindent
\textbf{Structure from motion.} We show per-scene decomposition of SfM on Tanks\&Temples~\cite{Knapitsch2017tankandtemple} in Tab.~\ref{tab:sup_sfm_tnt}. The COLMAP results~\cite{Riegler2020FVS} are used as the ground-truth.

\section{Qualitative examples}

\noindent
\textbf{Visual Odometry.} As in Fig.~\ref{fig:sup_vo_seven} and Fig.~\ref{fig:sup_vo_tum}, we present qualitative results of our visual odometry on all static scenes from the 7Scenes~\cite{shotton2013scorf}, TUM~\cite{sturm2012tumrgbd}, and ETH SLAM~\cite{schops2019badslam} datasets used in our evaluation.

\noindent
\textbf{Structure from Motion.} Fig.~\ref{fig:sup_sfm_eth} and Fig.~\ref{fig:sup_sfm_tnt} show qualitative results of SfM on all scenes from the ETH3D~\cite{schops2017eth3d} and Tanks\&Temples~\cite{Knapitsch2017tankandtemple} datasets. In Fig.~\ref{fig:sup_sfm_imc}, we further include qualitative results on the Cambridge Landmarks dataset, in-the-wild image collections, and the IMC Phototourism~\cite{jin2021imc} dataset, which contains large-scale unordered images captured by tourists.

\FloatBarrier
{
    \small
    \bibliographystyle{ieeenat_fullname}
    \bibliography{main}
}


\end{document}